\documentclass[10pt,twocolumn,letterpaper]{article}

\usepackage{cvpr}              
\usepackage{algorithm}
\usepackage{algorithmic}
\usepackage{multirow}
\usepackage{multicol}
\definecolor{cvprblue}{rgb}{0.21,0.49,0.74}
\usepackage[pagebackref,breaklinks,colorlinks,allcolors=cvprblue]{hyperref}

\def\paperID{9} 
\def\confName{CVPR}
\def\confYear{2026}

\title{Training for Trustworthy Saliency Maps: Adversarial Training Meets Feature-Map Smoothing}

\author{Dipkamal Bhusal\\
Rochester Institute of Technology\\ 
{\tt\small db1702@rit.edu}
\and
Md Tanvirul Alam\\
Rochester Institute of Technology\\
{\tt\small ma8235@rit.edu}
\and
Nidhi Rastogi\\
Rochester Institute of Technology\\
{\tt\small nxrvse@rit.edu}
}

\begin{document}
\maketitle
\begin{abstract}
Gradient-based saliency methods such as Vanilla Gradient (VG) and Integrated Gradients (IG) are widely used to explain image classifiers, yet the resulting maps are often noisy and unstable, limiting their usefulness in high-stakes settings. Most prior work improves explanations by modifying the attribution algorithm, leaving open how the training procedure shapes explanation quality. We take a training-centered view and first provide a curvature-based analysis linking attribution stability to how smoothly the input-gradient field varies locally. Guided by this connection, we study adversarial training and identify a consistent trade-off: it yields sparser and more input-stable saliency maps, but can degrade output-side stability, causing explanations to change even when predictions remain unchanged and logits vary only slightly. To mitigate this, we propose augmenting adversarial training with a lightweight feature-map smoothing block that applies a differentiable Gaussian filter in an intermediate layer. Across FMNIST, CIFAR-10, and ImageNette, our method preserves the sparsity benefits of adversarial training while improving both input-side stability and output-side stability. A human study with 65 participants further shows that smoothed adversarial saliency maps are perceived as more sufficient and trustworthy. Overall, our results demonstrate that explanation quality is critically shaped by training, and that simple smoothing with robust training provides a practical path toward saliency maps that are both sparse and stable. Code is available at {\url{https://github.com/dipkamal/robustness_plus_smoothing}}.
\end{abstract}    
\section{Introduction}
Saliency maps are among the most widely used tools for interpreting image classifiers, providing pixel-level visualizations of which input regions most influence a model’s prediction. Gradient-based methods such as Vanilla Gradient (VG)~\cite{simonyan2013deep} and Integrated Gradients (IG)~\cite{sundararajan2017axiomatic} are especially popular because they are simple, model-agnostic, and computationally efficient. However, in practice these explanations often suffer from noise, instability, limited faithfulness, and vulnerability to manipulation~\cite{adebayo2018sanity, kindermans2019reliability, nie2018theoretical, ghorbani2019interpretation, zhang2020interpretable}. These shortcomings raise a central question: \emph{when can we trust gradient-based explanations?}

For saliency maps to be trustworthy, they should satisfy several desiderata:  
(1) \emph{Sparsity}—explanations should be concise and focus on a small set of discriminative pixels~\cite{chalasani2020concise};  
(2) \emph{Stability}—explanations should remain consistent under small perturbations~\cite{alvarez2018robustness}; and 
(3) \emph{Faithfulness}—highlighted features should reflect what the model actually uses to make its decision~\cite{rong22consistent}. Most prior work pursues these goals by proposing new attribution algorithms or post-hoc smoothing schemes~\cite{smilkov2017smoothgrad, bykov2022noisegrad, shrikumar2017learning, springenberg2014striving, bach2015pixel}. In contrast, we adopt a training-centered view and ask:

\begin{quote}
\emph{How does the training procedure shape the quality of standard gradient-based explanations?}
\end{quote}

We start by analyzing the stability of gradient attributions under input perturbations. Under a single-layer model assumption, we show that attribution variation is controlled by the curvature of the underlying score function: for VG and IG, small changes in the input can produce large changes in the explanation when the gradient field varies rapidly, i.e., when the local Jacobian is highly non-Lipschitz. This perspective provides a principled link between \emph{model sensitivity} and \emph{explanation stability}, and motivates why training procedures that reduce local sensitivity should improve stability.

A natural candidate is \emph{adversarial training}~\cite{goodfellow2015explaining, madry2017towards}, which explicitly enforces prediction consistency in a neighborhood around each training example. Empirically, we confirm prior observations that adversarially trained models tend to produce \emph{sparser} saliency maps~\cite{etmann2019connection, chalasani2020concise} and improve \emph{input-side stability} under Gaussian noise. However, we uncover previously under-quantified trade-off: while adversarial training makes attributions sharper, it can \emph{degrade output-side stability} \cite{agarwal2022rethinking}. 

To address this failure mode, we propose a simple training-time regularizer: \emph{feature-map smoothing}. We insert a lightweight, differentiable Gaussian filter in intermediate layers during adversarial training to suppress high-frequency activation fluctuations that propagate into the input-gradient field. Intuitively, smoothing reduces abrupt changes in internal representations and dampens oscillations of the Jacobian, improving stability of gradient-based attributions. Across datasets, we show that this intervention preserves the sparsity benefits of adversarial training while improving both input-side stability (SSIM) and output-side stability (ROS), without sacrificing faithfulness as measured by ROAD~\cite{rong22consistent}. A human study further confirms that the resulting explanations are perceived as more sufficient and trustworthy.

We evaluate our approach on FMNIST~\cite{xiao2017fashion}, CIFAR-10~\cite{krizhevsky2009learning}, and ImageNette~\cite{howardsmaller}, comparing naturally trained models, adversarially trained models, and adversarially trained models with Gaussian feature-map smoothing. Our results show a consistent pattern:  
\textit{(i)} adversarial training increases sparsity but can harm output-side stability, and  
\textit{(ii)} feature-map smoothing recovers stability while largely preserving sparsity and faithfulness of adversarial training.  

\begin{figure}[t]
    \centering
    \includegraphics[width=0.7\linewidth]{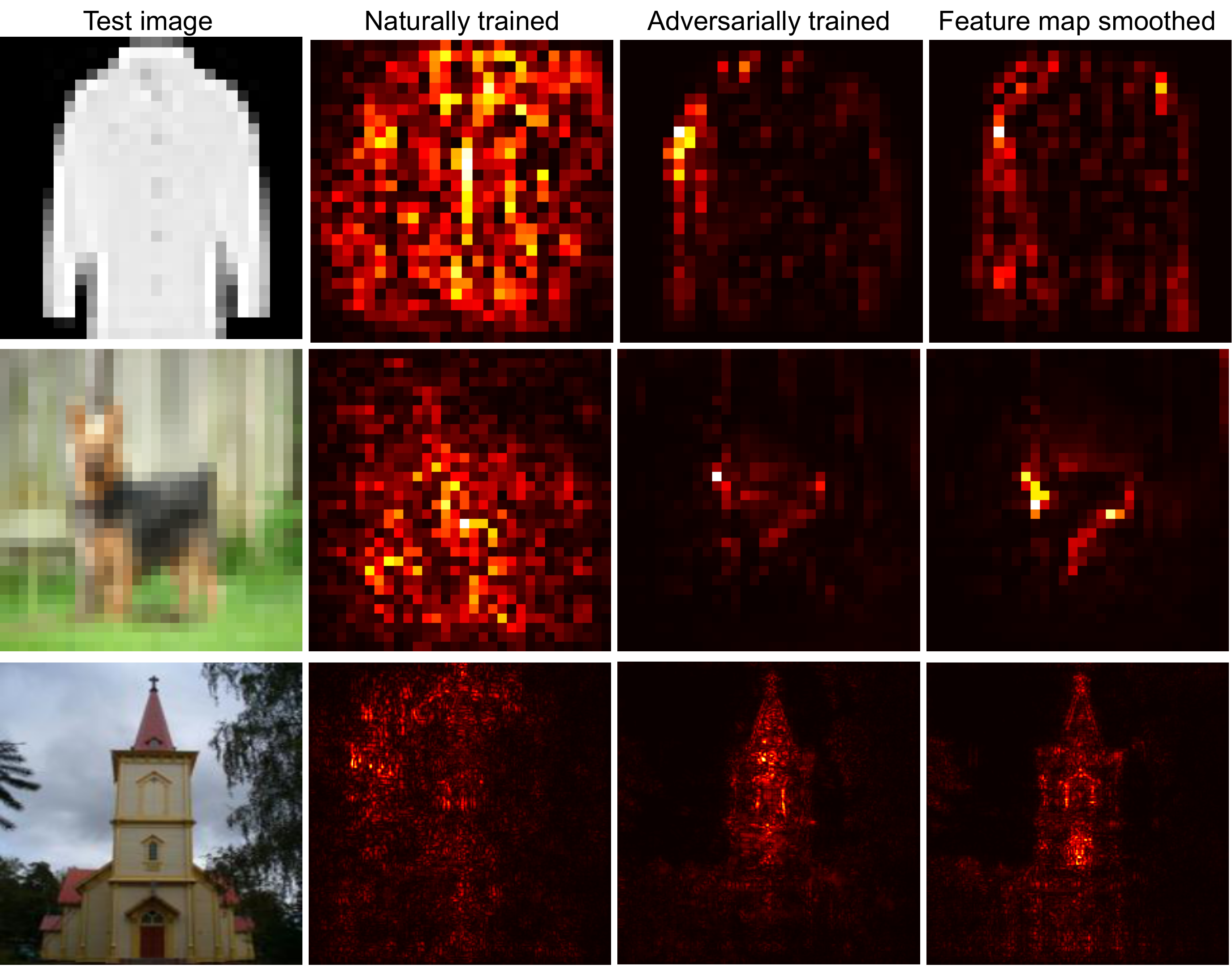}
    \caption{Saliency maps (Vanilla Gradient) for the same correctly classified input under three training regimes.
    \textbf{(a) Natural training:} noisy and diffuse maps that obscure the decision rationale.
    \textbf{(b) Adversarial training:} sparser maps but with reduced stability, sometimes discarding structurally relevant regions.
    \textbf{(c) Adversarial + Gaussian feature-map smoothing:} restores stability while retaining sparsity, producing explanations that are more coherent.}
    \label{fig:intro_example}
\end{figure}

Figure~\ref{fig:intro_example} illustrates the qualitative effect. Natural models produce noisy, diffuse maps; adversarial training yields sharper but sometimes brittle explanations; and adversarial training with feature-map smoothing produces saliency maps that are sparse and structurally coherent. We provide more examples in \cref{appendix:vizvisualizations}.

\textbf{Contributions.} Our main contributions are:
\begin{itemize}
    \item We provide a curvature-based analysis connecting the stability of Vanilla Gradient (VG)~\cite{simonyan2013deep} and Integrated Gradients (IG)~\cite{sundararajan2017axiomatic} saliency maps to the smoothness of the input-gradient field, motivating training-centered control of explanation stability.
    \item We empirically identify and quantify a trade-off induced by adversarial training: improved sparsity \cite{chalasani2020concise} and input-side stability \cite{adebayo2018sanity} can come at the cost of degraded output-side stability (ROS) \cite{agarwal2022rethinking}.
    \item We propose adversarial training with lightweight feature-map smoothing to mitigate this trade-off, improving stability while preserving sparsity, robustness, and faithfulness.
    \item We validate that these quantitative gains translate to human perception, showing that smoothed adversarial explanations are rated as more sufficient and trustworthy.
\end{itemize}

\section{Background}\label{sec:relatedwork}

\textbf{Feature attribution methods.} Feature attribution methods assign an importance score to each input feature for a given, visualized as saliency maps in images. Vanilla Gradient (VG)~\cite{simonyan2013deep} computes the gradient of the predicted-class score with respect to the input. While simple and efficient, VG produces noisy and visually cluttered maps. Integrated Gradients (IG)~\cite{sundararajan2017axiomatic} improves on VG by averaging gradients along a path from a baseline to the input, and satisfies axioms such as sensitivity and completeness~\cite{sturmfels2020visualizing}. SmoothGrad~\cite{smilkov2017smoothgrad} and NoiseGrad~\cite{bykov2022noisegrad} reduce noise by aggregation. Despite their popularity, gradient-based explanations can be sensitive to input perturbations, depend on hyperparameters, and can be manipulated without changing the predicted label~\cite{kindermans2019reliability, ghorbani2019interpretation, heo2019fooling, wagner2019interpretable}. This has motivated methods that add explanation-oriented regularization~\cite{kapishnikov2021guided, xu2020attribution, lim2021building, fong2017interpretable, yang2023idgi}.

\textbf{Training-centered approaches.} A growing body of work argues that explanation quality is shaped not only by the explainer, but also by the \emph{trained model} itself. Training-time interventions include gradient regularization~\cite{ross2017right}, explanation constraints~\cite{erion2021improving}, and architectural modifications that promote smoother representations~\cite{kim2019saliency, dombrowski2019explanations}. Adversarial training has received particular attention because robust models often exhibit visually sharper and sparser gradients. Etmann et al.~\cite{etmann2019connection} connect robustness to input-gradient alignment and interpretability on MNIST-scale settings. Zhang et al.~\cite{zhang2019interpreting} show that robust models tend to rely more on shape-biased representations, which can yield more structured saliency maps.
Chalasani et al.~\cite{chalasani2020concise} quantify sparsity improvements from $\ell_\infty$ adversarial training using the Gini index.

\section{Methodology}
\label{sec:method}

\subsection{Setup}
Let $z(\mathbf{x})\in\mathbb{R}^C$ denote the logits of a differentiable classifier, and
$\hat{y}(\mathbf{x})=\arg\max_k z_k(\mathbf{x})$ its predicted label.
For an input $\mathbf{x}$, we define the scalar score as the predicted-class logit
\begin{equation}
f(\mathbf{x}) := z_{\hat{y}(\mathbf{x})}(\mathbf{x}) \in \mathbb{R}.
\end{equation}

We consider two standard input-gradient attributions: \textbf{Vanilla Gradient (VG)}~\cite{simonyan2013deep} and \textbf{Integrated Gradients (IG)}~\cite{sundararajan2017axiomatic}. For \textbf{VG}, the feature attribution vector $\phi(\mathbf{x})$for an input $\mathbf{x}$ is given by:
\begin{equation}
\phi_{\mathrm{VG}}(\mathbf{x}) = \nabla_{\mathbf{x}} f(\mathbf{x}),
\label{eq:vg_def}
\end{equation}
which measures the sensitivity of the score to each input feature.

For \textbf{IG} with baseline $\mathbf{u}$:
\begin{equation}
\phi_{\mathrm{IG}}(\mathbf{x};\mathbf{u})
= (\mathbf{x}-\mathbf{u}) \odot \int_{0}^{1} \nabla_{\mathbf{x}} f\!\left(\mathbf{u}+\alpha(\mathbf{x}-\mathbf{u})\right)\,d\alpha,
\label{eq:ig_def}
\end{equation}
where $\odot$ is the Hadamard product. In practice, we approximate the integral with $S$ steps and use a zero baseline.

To formalize stability, we consider $\mathbf{x}' \in \mathcal{N}_\mathbf{x}$ a perturbed version of input image $\mathbf{x}$ where $\mathcal{N}_\mathbf{x}$ indicates a neighborhood of inputs $\mathbf{x}$ where the model prediction is locally consistent. We use $\|\cdot\|_2$ unless stated otherwise.

\subsection{A curvature view of attribution stability}
\label{sec:curvature_view}

For simplicity, let us examine a single-layer model with the following form:
$$
{f}(\mathbf{x}) = H(\langle \mathbf{w}, \mathbf{x} \rangle),
$$

where \( H \) is a differentiable scalar-valued activation function, \( \langle \mathbf{w}, \mathbf{x} \rangle \) is the dot product between the weight vector \( \mathbf{w} \) and input \( \mathbf{x} \in \mathbb{R}^d \).

Then, we obtain the feature attribution for Vanilla Gradients (VG), as:

\begin{equation}
    \phi_{\text{VG}}(\mathbf{x}) = H'(\langle \mathbf{w}, \mathbf{x} \rangle)\,\mathbf{w}.
\end{equation}

For $\mathbf{x}'\in \mathcal{N}_{\mathbf{x}}$:

\begin{equation}
    \phi_{\text{VG}}(\mathbf{x'}) = H'(\langle \mathbf{w}, \mathbf{x'} \rangle)\,\mathbf{w}.
\end{equation}

Let $s=\langle \mathbf{w},\mathbf{x}\rangle$ and $s'=\langle \mathbf{w},\mathbf{x}'\rangle$. Then, $\phi_{\text{VG}}(\mathbf{x}) = H'(s)\,\mathbf{w}$ and $\phi_{\text{VG}}(\mathbf{x}') = H'(s')\,\mathbf{w}$. Hence, the stability of explanations can be computed by measuring the norm of the difference between the perturbed-input explanation and the original explanation:
\begin{align}
\big\|\phi_{\text{VG}}(\mathbf{x}')-\phi_{\text{VG}}(\mathbf{x})\big\|
&= \big\|\big(H'(s')-H'(s)\big)\,\mathbf{w}\big\|
\\
&= \big|H'(s')-H'(s)\big|\,\|\mathbf{w}\|. 
\label{eq:vg-step1}
\end{align}

By the mean-value theorem applied to the scalar function $H'$ on the interval between $s$ and $s'$, there exists
$\xi$ between $s$ and $s'$ such that
\begin{align}
H'(s')-H'(s) &= H''(\xi)\,(s'-s).
\end{align}

\begin{figure*}[t]
    \centering
    \includegraphics[width=0.9\linewidth]{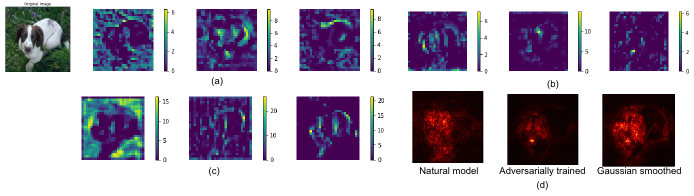}
    \caption{Effect of training regimes on intermediate feature maps (channel= \{7, 21, 127\}, after first residual block) and saliency maps:
    (a) naturally trained, (b) adversarially trained, (c) adversarially trained with feature-map smoothing, and (d) corresponding VG saliency maps.}
    \label{fig:imagenetfeaturemap_re}
\end{figure*}

Taking absolute values and using $s'-s=\langle \mathbf{w},\mathbf{x}'-\mathbf{x}\rangle$, we get
\begin{align}
\big|H'(s')-H'(s)\big|
&= \big|H''(\xi)\big|\,\big|\langle \mathbf{w},\mathbf{x}'-\mathbf{x}\rangle\big|
\\
&\le \Big(\sup_{z}\big|H''(z)\big|\Big)\,\big|\langle \mathbf{w},\mathbf{x}'-\mathbf{x}\rangle\big|.
\label{eq:vg-step2}
\end{align}

Applying Cauchy–Schwarz~\cite{aldaz2015advances},
\begin{align}
\big|\langle \mathbf{w},\mathbf{x}'-\mathbf{x}\rangle\big|
\;\le\; \|\mathbf{w}\|\,\|\mathbf{x}'-\mathbf{x}\|.
\label{eq:vg-step3}
\end{align}
Combining \eqref{eq:vg-step1}–\eqref{eq:vg-step3} yields
\begin{align}\label{eqn:VGstability}
\big\|\phi_{\text{VG}}(\mathbf{x}')-\phi_{\text{VG}}(\mathbf{x})\big\|
\;\le\; \Big(\sup_{z}\big|H''(z)\big|\Big)\,\|\mathbf{w}\|^2\,\|\mathbf{x}'-\mathbf{x}\|.
\end{align}
\noindent
Thus, the \textbf{VG} attribution stability is controlled by the activation curvature $\sup_z |H''(z)|$ and the weight norm $\|\mathbf{w}\|$.

For Integrated Gradients (IG), we use the closed-form expression for a single-layer model $f(\mathbf{x})=H(\langle \mathbf{w},\mathbf{x}\rangle)$~\cite{chalasani2020concise}:

\begin{equation}\label{eqn:singlelayerIGstatement}
\phi_{\text{IG}}(\mathbf{x};\mathbf{u})
= \big[H(\langle \mathbf{w}, \mathbf{x}\rangle)-H(\langle \mathbf{w}, \mathbf{u}\rangle)\big]\;
\frac{(\mathbf{x}-\mathbf{u})\odot \mathbf{w}}{\langle \mathbf{w}, \mathbf{x}-\mathbf{u}\rangle},
\end{equation}

{Proof for Eqn. \ref{eqn:singlelayerIGstatement} is provided in \cref{appendix:closedform}.} The stability of the explanation is defined similar to VG as: 
$$\Delta_{\text{IG}}=\|\phi_{\text{IG}}(\mathbf{x}';\mathbf{u})-\phi_{\text{IG}}(\mathbf{x};\mathbf{u})\|$$

for a nearby $\mathbf{x}'\in\mathcal{N}_{\mathbf{x}}$ with the same predicted label. By mean-value theorem and Cauchy–Schwarz~\cite{aldaz2015advances} yields
\begin{equation}\label{eqn:stabilityBoundIG}
    \begin{split}
\Delta_{\text{IG}}
\;\le\;
\Big(\underbrace{\sup_{z}\!|H'(z)|}_{\text{activation slope}}\;\|\mathbf{w}\| \\
\;+\;
\underbrace{\tfrac12\sup_{z}\!|H''(z)|}_{\text{activation curvature}}\;\|\mathbf{w}\|^2\,\|\mathbf{x}-\mathbf{u}\|\Big)
\;\|\mathbf{x}'-\mathbf{x}\|
\
    \end{split}
\end{equation}

Full derivation is provided in Appendix~\ref{appendix:IGfullproof}. Thus, like \textbf{VG}, \textbf{IG} stability is governed by the activation’s curvature and the weight scale; \textbf{IG} additionally depends (benignly) on the baseline distance $\|\mathbf{x}-\mathbf{u}\|$.

\paragraph{Takeaway.} \cref{eqn:VGstability} and \cref{eqn:stabilityBoundIG} shows that \emph{gradient explanations are stable when the model’s logit function has low effective curvature and the gradient field varies smoothly}.
This motivates focusing on \emph{training procedures} that shape local sensitivity and curvature of the predictor rather than designing new explainers.

\subsection{Adversarial training: invariance helps, but reveals a stability trade-off}
\label{sec:at_tradeoff}
A standard way to enforce local invariance is adversarial training~\cite{goodfellow2015explaining}, which solves the min--max objective:
\begin{equation}
\min_{\theta}\;\mathbb{E}_{(\mathbf{x},y)\sim \mathcal{D}}
\left[\max_{\|\boldsymbol{\delta}\|_\infty \le \epsilon} \mathcal{L}\!\left(\mathbf{x}+\boldsymbol{\delta},y;\theta\right)\right],
\label{eq:at_obj}
\end{equation}

where, the inner maximization finds the perturbation $\delta$ (bounded by $\epsilon$) that maximally increases the loss often approximated by projected gradient descent (PGD)~\cite{madry2017towards}, and the outer minimization updates the model to resist such worst-case perturbations. 

Because adversarial training explicitly enforces \emph{prediction consistency} over an $\ell_\infty$ ball, it often reduces input-level sensitivity and improves
\emph{input-side stability} (e.g., higher structural similarity under Gaussian input noise). Adversarial training also commonly produces \emph{sparser} saliency maps, concentrating attribution on fewer pixels~\cite{chalasani2020concise}. However, prediction invariance does not guarantee that the \emph{gradient field} is smooth in a way aligned with output behavior. Empirically, we observe a consistent trade-off:
\begin{center}
\emph{Adversarial training improves sparsity and input-side stability, but can worsen output-side stability (ROS),}
\end{center}
meaning that explanations may fluctuate even when logits change little and the predicted class is unchanged. This failure mode motivates adding a training-time prior that directly suppresses \emph{high-frequency representation changes} that amplify gradient oscillations.

\subsection{Feature-map smoothing as a stability prior}
\label{sec:smoothing}
We propose augmenting adversarial training with \emph{feature-map smoothing}~\cite{xie2019feature}, a lightweight architectural regularization that encourages smoother intermediate representations and, in turn, a smoother input--gradient field.

\begin{figure}[h]
    \centering
    \includegraphics[width=0.35\linewidth]{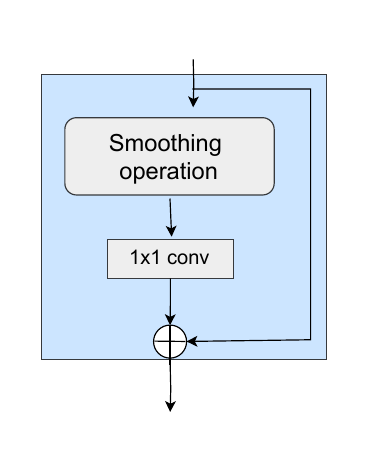}
    \caption{Feature-map smoothing block. A spatial filter is followed by a $1\times1$ convolution and a residual connection.}
    \label{fig:networkdenoisemap}
\end{figure}

Let $\mathbf{A}\in \mathbb{R}^{K\times H\times W}$ be a feature map with $K$ channels. We apply a differentiable spatial low-pass filter $\mathcal{G}_\sigma$ (Gaussian with bandwidth $\sigma$) independently per channel:
\begin{equation}
\mathbf{A}^{\mathrm{smooth}} = \mathcal{G}_\sigma(\mathbf{A}).
\end{equation}
We then use a $1\times1$ convolution to re-mix channels after spatial smoothing, and add a residual connection:
\begin{equation}
\mathrm{SmoothBlock}(\mathbf{A}) \;=\; \mathbf{A} \;+\; \mathrm{Conv}_{1\times1}\!\left(\mathbf{A}^{\mathrm{smooth}}\right).
\label{eq:smoothblock}
\end{equation}
Figure~\ref{fig:networkdenoisemap} illustrates the block. In practice, $\mathcal{G}_\sigma$ is implemented as a fixed depthwise convolution (per-channel Gaussian kernel), while $\mathrm{Conv}_{1\times1}$ is learned and preserves the channel dimension.

Using a spatial smoothing suppresses high-frequency activation components that can induce sharp, localized changes in $f(\mathbf{x})$ and its derivatives.
Intuitively, this reduces the effective curvature of the end-to-end mapping and stabilizes $\nabla_{\mathbf{x}} f(\mathbf{x})$,
while the residual path preserves representational capacity and model accuracy.

\paragraph{Training with smoothing.}
Our final method performs adversarial training \cref{eq:at_obj} on a network where SmoothBlocks are inserted at selected intermediate layers. Unless stated otherwise, we insert the SmoothBlock after the first convolutional/residual block (validated in \cref{ablation:position}). This preserves the sparsity and invariance benefits of adversarial training, while empirically restoring stability without sacrificing sparsity. \cref{fig:imagenetfeaturemap_re}(c–d) illustrates smoothing effect in the intermediate feature maps: it reduces high-frequency noise, producing saliency maps that are sparse yet structurally coherent.

\subsection{Algorithm}
\label{sec:algo}
Algorithm~\ref{alg:at_smooth} summarizes training. We generate PGD adversarial examples and update parameters using the adversarial loss.
Smoothing is implemented as part of the forward pass through the SmoothBlocks.

\begin{algorithm}[t]
\caption{Adversarial training with feature-map smoothing}
\label{alg:at_smooth}
\begin{algorithmic}[1]
\REQUIRE Training data $\mathcal{D}$, smoothed model $f_\theta$ (SmoothBlocks in forward pass), PGD budget $\epsilon$, steps $T$, step size $\alpha$
\FOR{each minibatch $(\mathbf{x},y)$}
    \STATE Initialize $\boldsymbol{\delta}_0 \sim \mathcal{U}([-\epsilon,\epsilon])$ (optional random start)
    \FOR{$t=0$ to $T-1$}
        \STATE $\mathbf{x}_t \leftarrow \mathrm{clip}(\mathbf{x}+\boldsymbol{\delta}_t,\;0,\;1)$ \COMMENT{project to valid input range}
        \STATE $\mathbf{g}_t \leftarrow \nabla_{\boldsymbol{\delta}} \mathcal{L}\!\left(f_\theta(\mathbf{x}_t),y\right)$ \COMMENT{grad flows through SmoothBlocks}
        \STATE $\boldsymbol{\delta}_{t+1} \leftarrow \Pi_{\|\boldsymbol{\delta}\|_\infty \le \epsilon}\!\left(\boldsymbol{\delta}_{t} + \alpha\cdot \mathrm{sign}(\mathbf{g}_t)\right)$
    \ENDFOR
    \STATE $\mathbf{x}_{adv} \leftarrow \mathrm{clip}(\mathbf{x}+\boldsymbol{\delta}_{T},\;0,\;1)$
    \STATE Update $\theta$ by minimizing $\mathcal{L}\!\left(f_\theta(\mathbf{x}_{adv}),y\right)$
\ENDFOR
\end{algorithmic}
\end{algorithm}

\paragraph{Implementation notes.}
We use Gaussian filters for $\mathcal{G}_\sigma$ due to differentiability and well-understood smoothing behavior, and include mean/median filter ablations in Sec.~\ref{ablation:meanMedian}.

\section{Experiment and Analysis}\label{sec:experimentAndAnalysis}

\paragraph{Setup.}
We evaluate saliency-map quality on three image classification benchmarks: FMNIST~\cite{xiao2017fashion}, CIFAR-10~\cite{krizhevsky2009learning} and ImageNette \cite{howardsmaller}. We train three model variants per dataset:
\textbf{(N)} naturally trained, \textbf{(A)} adversarially trained, and \textbf{(G)} adversarially trained with an additional local \emph{Gaussian feature-map smoothing} block. We use LeNet \cite{lecun1998gradient} for FMNIST and ResNet \cite{he2016deep} for CIFAR-10 and ImageNette (additional results with VGG-16~\cite{simonyan2014very} in \cref{appendix:additionalexperiments}).  The hyper-parameters of PGD attack in our adversarial training: for FMNIST and CIFAR-10, $\epsilon$ = 0.1, attack step size = $\epsilon/10$, and number of iterations = 40; for ImageNette  $\epsilon$ = 1/255, step size = 0.00784 and number of iterations = 20. The smoothing block is inserted after the first convolutional/residual block (validated in \cref{ablation:position}). Unless stated otherwise, all training hyperparameters are kept identical across \textbf{N}, \textbf{A}, and \textbf{G} so that differences in explanation quality can be attributed to the training procedure rather than optimization choices (training details in \cref{appendix:training}).

\paragraph{Evaluation metrics.}
For each dataset and model variant, we compute saliency maps using Vanilla Gradient (VG) and Integrated Gradients (IG).
We evaluate {sparsity} via Gini index \cite{chalasani2020concise},
{input-side stability} via SSIM under Gaussian input noise \cite{adebayo2018sanity},
{output-side stability} via Relative Output Stability  \cite{agarwal2022rethinking},
and {faithfulness} via ROAD-AOPC \cite{rong22consistent}. All reported metrics are averaged over 1000 randomly selected test images that are correctly classified \emph{by all three} model variants. Metric definitions are provided in \cref{appendix:metrics}.

\subsection{Results and Discussion}

\paragraph{Model performance.}
Before analyzing explanation quality, we verify that our training interventions do not catastrophically alter predictive performance.
Table~\ref{tab:my-table} reports standard (natural) accuracy and PGD robust accuracy for \textbf{N}, \textbf{A}, and \textbf{G}.
As expected, adversarial training (\textbf{A}) substantially increases robust accuracy at the cost of lower natural accuracy, reflecting the standard robustness--accuracy trade-off.
Crucially, incorporating Gaussian smoothing (\textbf{G}) largely preserves the robustness of \textbf{A} and changes natural accuracy by small margins on FMNIST and CIFAR-10 (and moderately on ImageNette).

\begin{table}[h]
\centering
\caption{Model performance on natural (N), adversarial (A) and Gaussian-smoothed adversarial (G) models.}
\label{tab:my-table}
\resizebox{0.4\textwidth}{!}{%
\begin{tabular}{@{}lcccccc@{}}
\toprule
\textbf{}           & \multicolumn{3}{c}{\textbf{Natural accuracy \%}} & \multicolumn{3}{c}{\textbf{Robust accuracy \%}} \\ \midrule
\textbf{} & \multicolumn{1}{c}{N} & \multicolumn{1}{c}{A} & \multicolumn{1}{c|}{G} & \multicolumn{1}{c}{N} & \multicolumn{1}{c}{A} & \multicolumn{1}{c}{G} \\ \midrule
\textbf{FMNIST}     & 89.9   & 79.9   & \multicolumn{1}{c|}{80.3}   & 9.5          & 67.7          & 66.8          \\
\textbf{CIFAR-10}   & 90.9   & 80.5   & \multicolumn{1}{c|}{80.8}   & 4.8          & 54.3          & 53.9          \\
\textbf{ImageNette} & 96.3   & 70.8   & \multicolumn{1}{c|}{66.6}   & 1.6          & 12.2          & 13.5          \\ \bottomrule
\end{tabular}%
}
\end{table}

\paragraph{Sparsity vs. output-side stability reveals a trade-off.}
Table~\ref{tab:sparsity-stability} summarizes sparsity and output-side stability for VG and IG. Consistent with prior work \cite{chalasani2020concise}, adversarial training (\textbf{A}) increases sparsity across datasets and attribution methods.
However, this often comes with \emph{worse} ROS.
This exposes a previously under-quantified trade-off that
\emph{adversarial training improves sparsity, but can harm output-side stability (ROS).} Importantly, Gaussian smoothing (\textbf{G}) consistently improves ROS relative to \textbf{A} while preserving most of the sparsity gains.

\begin{table}[h]
\centering
\caption{Sparsity (Gini) and Relative Output Stability (ROS) for Vanilla Gradient (VG) and Integrated Gradients (IG) on FMNIST, CIFAR-10, and ImageNette models: \textbf{N} (naturally trained), \textbf{A} (adversarially trained) \& \textbf{G} (gaussian-filter smoothed adversarially trained). $\uparrow$ \& $\downarrow$ means higher \& lower values are better.}
\label{tab:sparsity-stability}
\resizebox{0.45\textwidth}{!}{
\begin{tabular}{@{}llccc|ccc|ccc@{}}
\toprule
& & \multicolumn{3}{c}{\textbf{FMNIST}} & \multicolumn{3}{c}{\textbf{CIFAR-10}} & \multicolumn{3}{c}{\textbf{ImageNette}} \\
& & \textbf{N} & \textbf{A} & \textbf{G} & \textbf{N} & \textbf{A} & \textbf{G} & \textbf{N} & \textbf{A} & \textbf{G} \\
\midrule
\multirow{2}{*}{\textbf{VG}}
& \textbf{Gini} $\uparrow$ & 0.54 & {0.74} & 0.73 & 0.49 & {0.68} & 0.67 & 0.51 & 0.53 & {0.54} \\
& \textbf{ROS} $\downarrow$ & {11.03} & 13.12 & 11.77 & {2.33} & 2.55 & 2.54 & {2.74} & 2.88 & 2.77 \\
\midrule
\multirow{2}{*}{\textbf{IG}}
& \textbf{Gini} $\uparrow$ & 0.77 & {0.83} & 0.82 & 0.61 & {0.71} & {0.71} & 0.63 & 0.66 & {0.67} \\
& \textbf{ROS} $\downarrow$ & {7.26} & 9.19 & 8.56 & {4.25} & 4.48 & 4.36 & {5.56} & 5.60 & 5.57 \\
\bottomrule
\end{tabular}
}
\end{table}

\begin{figure}[h!]
    \centering
    \begin{subfigure}{0.5\textwidth}
        \centering
        \begin{subfigure}[b]{.45\textwidth}
            \centering
            \includegraphics[width=0.9\linewidth]{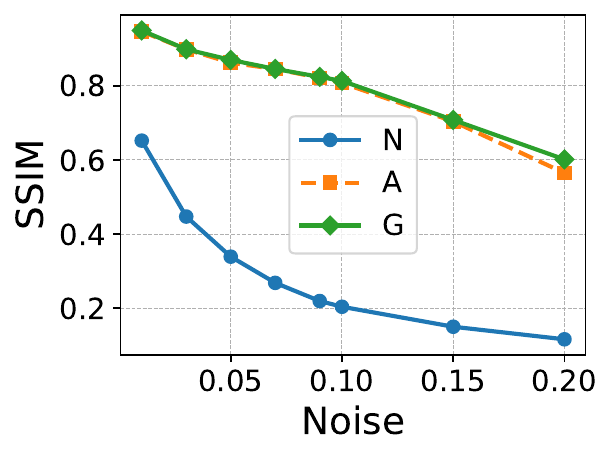}
            \caption{Vanilla Gradient (VG)}
            \label{fig:VGFMNISTSSIM_combined}
        \end{subfigure}
        \hfill
        \begin{subfigure}[b]{.45\textwidth}
            \centering
            \includegraphics[width=0.9\linewidth]{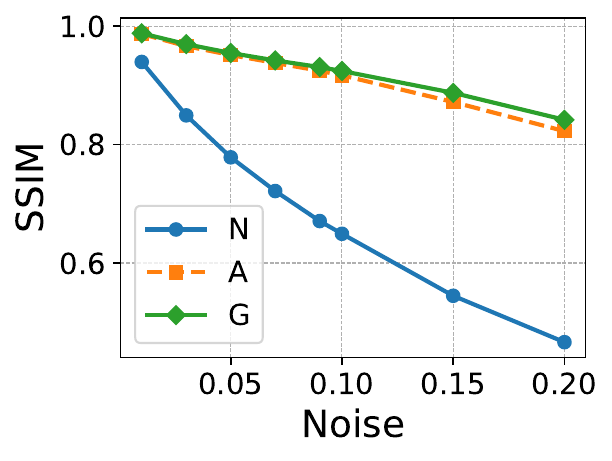}
            \caption{Integrated Gradients (IG)}
            \label{fig:IGFMNISTSSIM_combined}
        \end{subfigure}
        \caption{FMNIST}
        \label{fig:FMNISTSSIM_row}
    \end{subfigure}

    \vspace{0.5cm} 

    \begin{subfigure}{0.5\textwidth}
        \centering
        \begin{subfigure}[b]{.45\textwidth}
            \centering
            \includegraphics[width=0.9\linewidth]{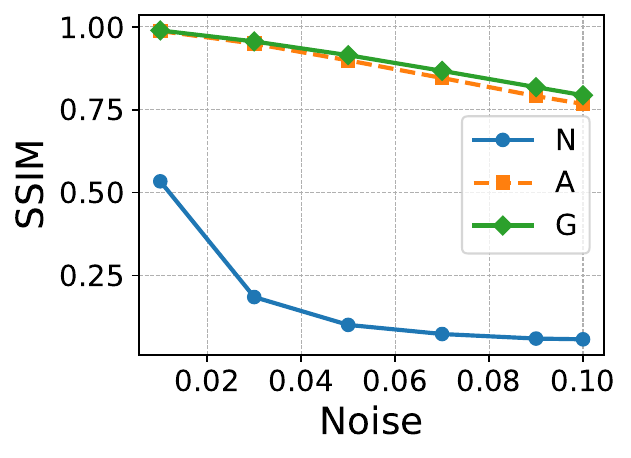}
            \caption{Vanilla Gradient (VG)}
            \label{fig:VGCIFARSSIM_combined}
        \end{subfigure}
        \hfill
        \begin{subfigure}[b]{.45\textwidth}
            \centering
            \includegraphics[width=0.9\linewidth]{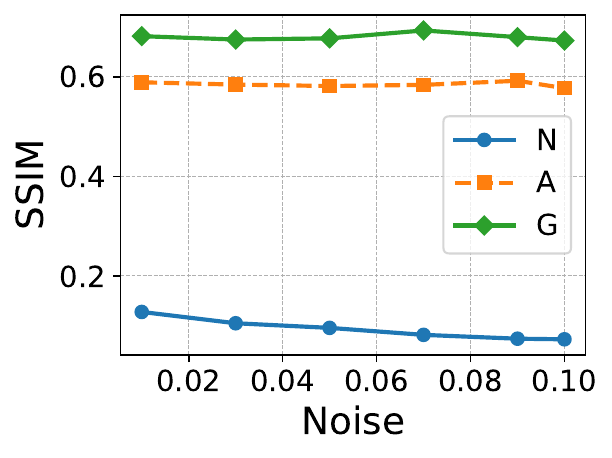}
            \caption{Integrated Gradients (IG)}
            \label{fig:IGCIFARSSIM_combined}
        \end{subfigure}
        \caption{CIFAR-10}
        \label{fig:CIFARSSIM_row}
    \end{subfigure}

    \vspace{0.5cm} 

    \begin{subfigure}{0.5\textwidth}
        \centering
        \begin{subfigure}[b]{.45\textwidth}
            \centering
            \includegraphics[width=0.9\linewidth]{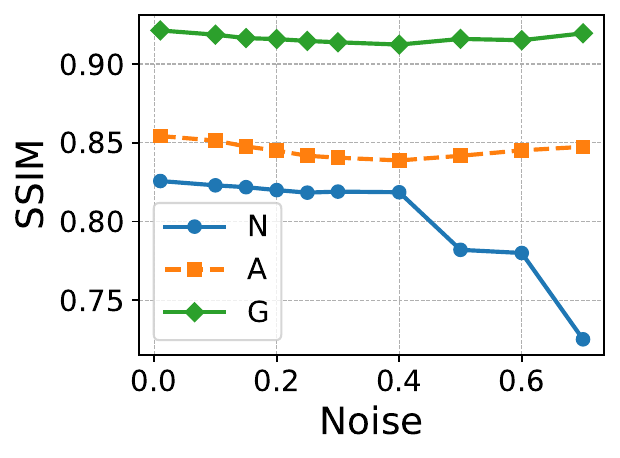}
            \caption{Vanilla Gradient (VG)}
            \label{fig:VGImageNetSSIM_combined}
        \end{subfigure}
        \hfill
        \begin{subfigure}[b]{.45\textwidth}
            \centering
            \includegraphics[width=0.9\linewidth]{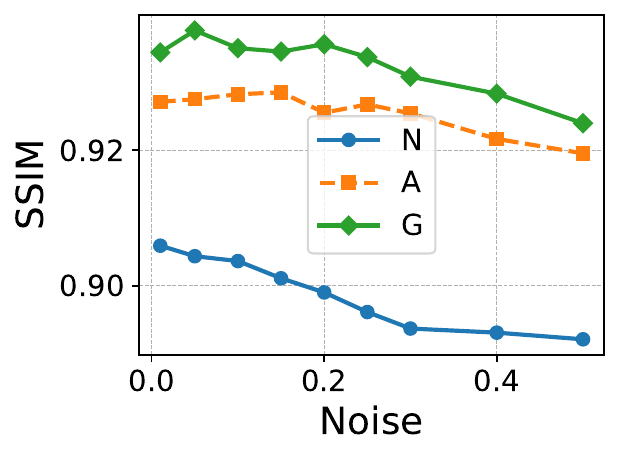}
            \caption{Integrated Gradients (IG)}
            \label{fig:IGImageNetSSIM_combined}
        \end{subfigure}
        \caption{ImageNette}
        \label{fig:ImageNetteSSIM_row}
    \end{subfigure}

    \caption{Structural similarity (SSIM) evaluation of saliency maps for naturally trained (N), adversarially trained (A), and adversarially trained with Gaussian smoothing (G).}
    \label{fig:combined_ssim_evaluation}
\end{figure}

\paragraph{Input-side stability improves with adversarial training and further with smoothing.}
Figure~\ref{fig:combined_ssim_evaluation} demonstrates input-side stability plots using structural similarity (SSIM) between saliency maps of clean inputs and prediction-preserving Gaussian-noisy inputs.
Natural training (\textbf{N}) is consistently the least stable: saliency maps degrade rapidly with increasing noise.
Adversarial training (\textbf{A}) improves SSIM across datasets, aligning with the view that enforcing local invariance suppresses noisy saliency patterns.
Adding Gaussian smoothing (\textbf{G}) further improves SSIM, especially at moderate-to-high noise levels, and the effect is most pronounced on ImageNette.

\paragraph{Faithfulness is preserved under smoothing.}
We evaluate faithfulness using ROAD-AOPC \cite{rong22consistent}, where higher AOPC indicates that the explanation correctly identifies pixels that drive the prediction.
Table~\ref{tab:faithfulness} reports ROAD-AOPC for VG and IG across all training regimes.
Adversarially trained models (\textbf{A}) typically achieve higher AOPC than natural models (\textbf{N}).
Gaussian smoothing (\textbf{G}) maintains AOPC on par with (and sometimes better than) \textbf{A}, indicating that the stability improvements from smoothing do not come at the expense of faithfulness.
Associated ROAD curves are shown in \cref{fig:combined_road_plot}, where a faster accuracy drop under MoRF removal corresponds to more faithful attributions.

\begin{table}[h]
\centering
\caption{ROAD AOPC (higher $\uparrow$ is more faithful) for Vanilla Gradient (VG) and Integrated Gradients (IG) on different models of FMNIST, CIFAR-10, and ImageNette: \textbf{N}: naturally trained, \textbf{A}: adversarially trained \& \textbf{G}: adversarially trained with Gaussian smoothing.}
\label{tab:faithfulness}
\resizebox{0.40\textwidth}{!}{%
\begin{tabular}{@{}lccc|ccc|ccc@{}}
\toprule
 & \multicolumn{3}{c|}{\textbf{FMNIST}} & \multicolumn{3}{c|}{\textbf{CIFAR-10}} & \multicolumn{3}{c}{\textbf{ImageNette}} \\
 & \textbf{N} & \textbf{A} & \textbf{G} & \textbf{N} & \textbf{A} & \textbf{G} & \textbf{N} & \textbf{A} & \textbf{G} \\ \midrule
\textbf{VG} & 0.22 & 0.21 & 0.23 & 0.20 & 0.23 & 0.24 & 0.13 & 0.13 & 0.19 \\
\textbf{IG} & 0.32 & 0.31 & 0.28 & 0.42 & 0.49 & 0.49 & 0.33 & 0.40 & 0.38 \\
\bottomrule
\end{tabular}}
\end{table}

\section{Human Perception of Explanation Quality}\label{sec:humanstudy}

Our quantitative results show that adversarial training increases saliency sparsity and improves input-side stability, but can degrade output-side stability (ROS), while adding Gaussian feature-map smoothing restores stability without sacrificing sparsity. We next ask whether these quantitative changes translate into improvements in \emph{human-perceived} explanation quality.

\subsection{Motivation}
Saliency maps are ultimately used by humans to reason about model decisions. Functionally grounded metrics (e.g., sparsity \cite{chalasani2020concise}, SSIM \cite{adebayo2018sanity}, ROS \cite{agarwal2022rethinking}, ROAD \cite{rong22consistent}) provide useful proxies, but they do not directly measure whether an explanation is perceived as understandable. We therefore ask:

\begin{quote}
\textit{Do the stability improvements from feature-map smoothing make gradient-based saliency maps more sufficient and trustworthy to humans?}
\end{quote}

\subsection{Study design}
We conducted a user study with 65 graduate students (Ph.D. and M.S.) with at least one year of experience in computer vision\footnote{The study was IRB-approved by our institution.}. Participants were shown saliency maps generated using Vanilla Gradients (VG) for 10 randomly selected test images drawn from FMNIST and CIFAR-10. For each image, we generated VG saliency maps from three models: naturally trained (\textbf{N}), adversarially trained (\textbf{A}), and adversarially trained with Gaussian smoothing (\textbf{G}). Thus, each participant evaluated $20 \times 3 = 60$ image--saliency pairs. Trials were presented in randomized order and model identity was not revealed (see~\cref{fig:survey1} in \cref{appendix:userstudy}).

For each trial, participants rated the explanation on a 5-point Likert scale (1=strongly disagree, 5=strongly agree) along two axes adapted from the Hoffman satisfaction scale~\cite{hoffman2023measures}:
\begin{itemize}
    \item \textbf{Sufficiency:} ``The explanation provides sufficient information to understand the model’s decision.''
    \item \textbf{Trust:} ``Based on this explanation, I trust the model’s classification.''
\end{itemize}

In a final block, participants were shown side-by-side saliency maps (\textbf{N}, \textbf{A}, \textbf{G}) for the same input and selected the most comprehensible explanation, providing brief justifications (see~\cref{fig:survey2} in \cref{appendix:userstudy}).
\vspace{-3mm}
\subsection{Results}
Table~\ref{tab:humanstudy} summarizes the mean and standard deviation of participant ratings. Overall, \textbf{G} received the highest ratings for both sufficiency and trust, followed by \textbf{A}, with \textbf{N} consistently lowest. Statistical tests confirm significance: Wilcoxon signed-rank tests across all pairs yielded $p < 10^{-10}$, and one-way ANOVA showed highly significant group effects (F-stat $>$ 190, $p < 10^{-70}$) (see ~\cref{tab:stattest} in \cref{appendix:userstudy}). 

\begin{table}[h]
\centering
\caption{Human study ratings (mean $\pm$ std). Higher is better.}
\label{tab:humanstudy}
\resizebox{0.42\textwidth}{!}{%
\begin{tabular}{lccc}
\toprule
 & Natural (N) & Adv (A) & Adv + Gaussian (G) \\
\midrule
Sufficiency & 2.08 $\pm$ 0.75 & 2.99 $\pm$ 0.93 & \textbf{3.33 $\pm$ 1.03} \\
Trust       & 2.02 $\pm$ 0.82 & 3.08 $\pm$ 0.90 & \textbf{3.14 $\pm$ 1.01} \\
\bottomrule
\end{tabular}%
}
\end{table}

In the side-by-side preference task, 58\% of participants selected \textbf{G} as the most comprehensible explanation, 27\% selected \textbf{A}, and 15\% selected \textbf{N}.
Free-text responses aligned with these preferences.
Naturally trained maps were frequently described as \textit{``too noisy''} or \textit{``highlighting irrelevant pixels.''}
Adversarially trained maps were often described as \textit{``sharper but incomplete,''} with comments that they \textit{``miss structure around the object.''}
Gaussian-smoothed adversarial maps were most commonly praised for clarity and alignment, e.g., \textit{``highlights important features without excessive noise''} and \textit{``better matches the object.''}

\section{Ablation studies}\label{sec:ablations}



\subsection{Position of the smoothing block}\label{ablation:position}
We study where feature-map smoothing is most effective.
Using adversarially trained CIFAR-10 models, we insert the smoothing block after the first, second, or third residual block and evaluate saliency quality with Vanilla Gradient (VG).
Table~\ref{tab:position-ablation} reports changes in sparsity ($\Delta$Gini, higher is better) and output-side stability ($\Delta$ROS, lower is better) relative to natural training.

\begin{table}[h]
\centering
\caption{\textbf{Ablation:} Effect of Gaussian smoothing block placement on CIFAR-10: changes in sparsity ($\Delta$Gini $\uparrow$) and Relative Output Stability ($\Delta$ ROS $\downarrow$) relative to natural training.}
\label{tab:position-ablation}
\resizebox{0.35\textwidth}{!}{%
\begin{tabular}{@{}lcc@{}}
\toprule
\multicolumn{1}{c}{\textbf{Placement}} & \textbf{$\Delta$Gini $\uparrow$} & \textbf{$\Delta$ ROS $\downarrow$} \\ \midrule
\textbf{After first block}             & 0.18               & 0.21             \\
\textbf{After second block}            & 0.18               & 0.23             \\
\textbf{After third block}             & 0.19               & 0.24             \\ \bottomrule
\end{tabular}%
}
\end{table}

Across placements, sparsity gains are similar ($\Delta$Gini $\approx +0.18$ to $+0.19$), with slightly higher sparsity when smoothing is applied deeper.
In contrast, stability improvements are strongest when smoothing is applied earlier (after the first block).
To balance sparsity and stability, we insert the smoothing block after the first residual block in all main experiments.

\subsection{Mean and median filters}\label{ablation:meanMedian}
Our main experiments focus on Gaussian smoothing due to its differentiability and smooth attenuation of high-frequency components.
To test whether improvements are specific to Gaussian filtering or reflect a more general \emph{local smoothing} principle, we replace the Gaussian filter with mean (\textbf{M1}) and median (\textbf{M2}) filters and re-evaluate sparsity (Gini), output-side stability (ROS), and input-side stability (SSIM).\footnote{We use differentiable local filtering operators implemented in Kornia~\cite{eriba2020kornia}.}

\begin{table}[h]
\centering
\caption{\textbf{Ablation}: Sparsity (Gini) and Relative Output Stability (ROS) for Vanilla Gradient (VG) and Integrated Gradients (IG) on models of FMNIST, CIFAR-10, \& ImageNette: \textbf{N}: naturally trained, \textbf{A}: adversarially trained, \textbf{G}: adversarially trained with Gaussian smoothing, \textbf{M1}: adversarially trained with Mean smoothing, \& \textbf{M2}: adversarially trained with Median smoothing.}
\label{tab:ablation-mean-median}
\resizebox{0.35\textwidth}{!}{%
\begin{tabular}{@{}ll|cc|cc@{}}
\toprule
                    &             & \multicolumn{2}{c|}{\textbf{VG}} & \multicolumn{2}{c}{\textbf{IG}} \\ \midrule
                    &             & \textbf{Gini} $\uparrow$     & \textbf{ROS} $\downarrow$   & \textbf{Gini} $\uparrow$    & \textbf{ROS} $\downarrow$    \\ \midrule
\textbf{FMNIST}     & \textbf{N}  & 0.54           & 11.03           & 0.77          & 7.26            \\
                    & \textbf{A}  & 0.74           & 13.12           & 0.83          & 9.19            \\
                    & \textbf{G}  & 0.73           & 11.77           & 0.82          & 8.56            \\
                    & \textbf{M1} & 0.74           & 12.16           & 0.84          & 10.18           \\
                    & \textbf{M2} & 0.72           & 10.01           & 0.81          & 11.94           \\ \midrule
\textbf{CIFAR-10}   & \textbf{N}  & 0.49           & 2.33            & 0.61          & 4.25            \\
                    & \textbf{A}  & 0.68           & 2.55            & 0.71          & 4.48            \\
                    & \textbf{G}  & 0.67           & 2.54            & 0.71          & 4.16            \\
                    & \textbf{M1} & 0.67           & 2.59            & 0.71          & 4.60            \\
                    & \textbf{M2} & 0.67           & 2.54            & 0.71          & 4.13            \\ \midrule
\textbf{ImageNette} & \textbf{N}  & 0.51           & 2.74            & 0.63          & 5.56            \\
                    & \textbf{A}  & 0.53           & 2.88            & 0.66          & 5.60            \\
                    & \textbf{G}  & 0.54           & 2.77            & 0.67          & 5.59            \\
                    & \textbf{M1} & 0.54           & 2.52            & 0.67          & 5.03            \\
                    & \textbf{M2} & 0.58           & 2.55            & 0.69          & 5.08            \\ \bottomrule
\end{tabular}%
}
\end{table}

\paragraph{Sparsity and output-side stability.}
Table~\ref{tab:ablation-mean-median} summarizes Gini and ROS for VG and IG across FMNIST, CIFAR-10, and ImageNette.
All smoothing variants (\textbf{G}, \textbf{M1}, \textbf{M2}) preserve the sparsity gains of adversarial training, with only small fluctuations in Gini ($\pm 0.01$--$0.02$).
ROS shows larger variation across filters:
\begin{itemize}
    \item \textbf{FMNIST:} Median smoothing (\textbf{M2}) yields the strongest ROS improvement for VG, while Gaussian smoothing (\textbf{G}) performs best for IG.
    \item \textbf{CIFAR-10:} Differences among Gaussain (G), Mean (M1), and Median (M2) smoothing are subtle.
    \item \textbf{ImageNette:} Mean and median filters reduce ROS more aggressively than Gaussian.
\end{itemize}

\begin{figure}[h]
    \centering

    \begin{subfigure}{0.5\textwidth}
        \centering
        \begin{subfigure}[b]{.45\textwidth}
            \centering
            \includegraphics[width=0.9\linewidth]{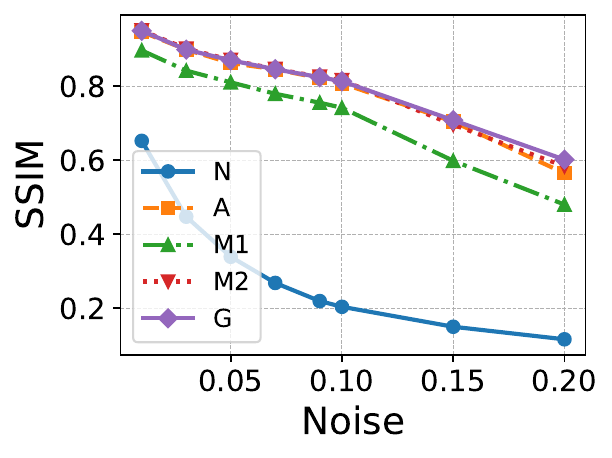}
            \caption{Vanilla Gradient (VG)}
            \label{fig:VGablationSSIMfmnist}
        \end{subfigure}
        \hfill
        \begin{subfigure}[b]{.45\textwidth}
            \centering
            \includegraphics[width=0.9\linewidth]{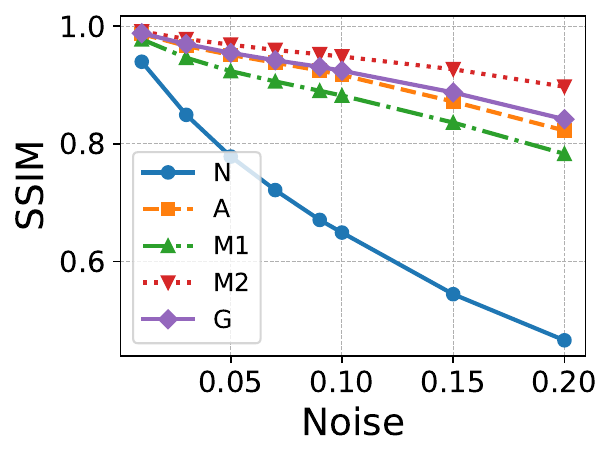}
            \caption{Integrated Gradients (IG)}
            \label{fig:IGablationSSIMfmnist}
        \end{subfigure}
        \caption{FMNIST}
        \label{fig:ablationSSIMfmnist}
    \end{subfigure}

    \vspace{0.5cm} 

    \begin{subfigure}{0.5\textwidth}
        \centering
        \begin{subfigure}[b]{.45\textwidth}
            \centering
            \includegraphics[width=0.9\linewidth]{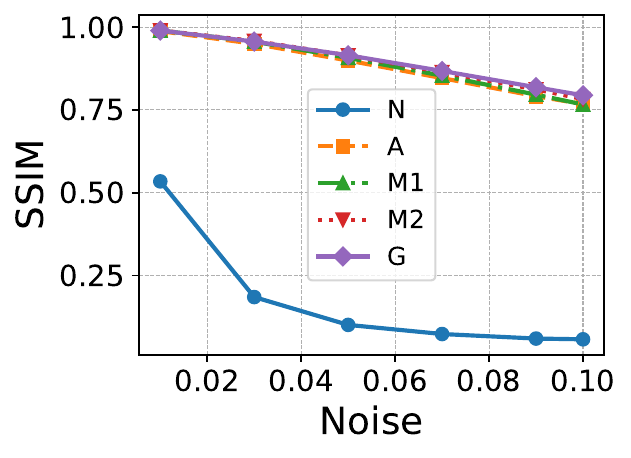}
            \caption{Vanilla Gradient (VG)}
            \label{fig:VGablationSSIMcifar}
        \end{subfigure}
        \hfill
        \begin{subfigure}[b]{.45\textwidth}
            \centering
            \includegraphics[width=0.9\linewidth]{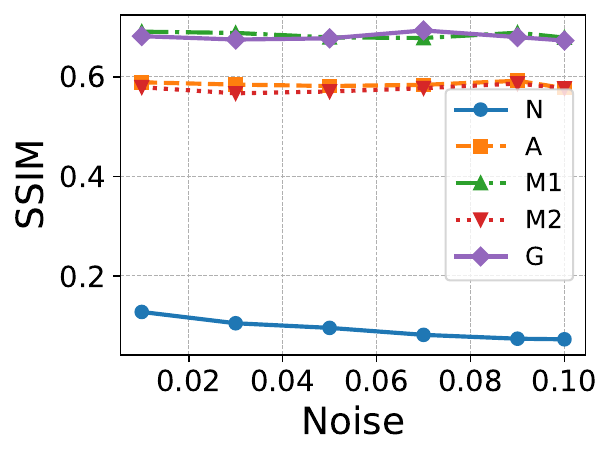}
            \caption{Integrated Gradients (IG)}
            \label{fig:IGablationSSIMcifar}
        \end{subfigure}
        \caption{CIFAR-10}
        \label{fig:ablationSSIMcifar}
    \end{subfigure}

    \vspace{0.5cm} 

    \begin{subfigure}{0.5\textwidth}
        \centering
        \begin{subfigure}[b]{.45\textwidth}
            \centering
            \includegraphics[width=0.9\linewidth]{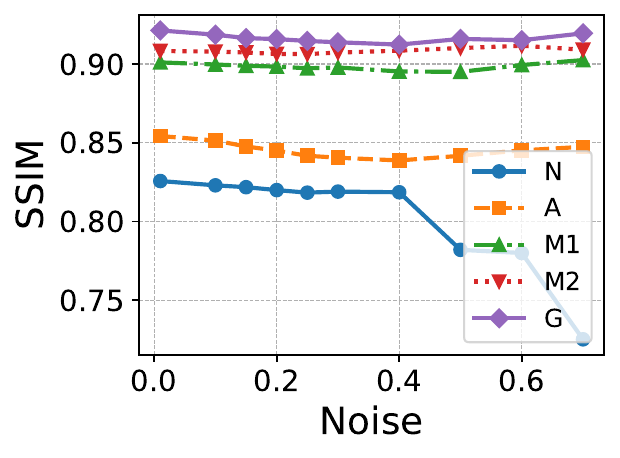}
            \caption{Vanilla Gradient (VG)}
            \label{fig:VGablationSSIMimagenette}
        \end{subfigure}
        \hfill
        \begin{subfigure}[b]{.45\textwidth}
            \centering
            \includegraphics[width=0.9\linewidth]{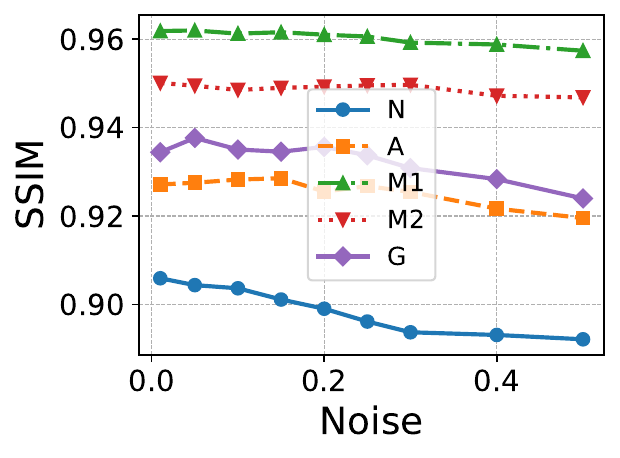}
            \caption{Integrated Gradients (IG)}
            \label{fig:IGablationSSIMimagenette}
        \end{subfigure}
        \caption{ImageNette}
        \label{fig:ablationSSIMimagenette}
    \end{subfigure}

    \caption{\textbf{Ablation:} Structural similarity evaluation of saliency maps on various ImageNette models: naturally trained \textbf{(N)}, adversarially trained \textbf{(A)}, and adversarial trained with smoothing filters (\textbf{M1}: mean filter, \textbf{M2}: median filter and \textbf{(G)}: Gaussian filter). }
    \label{fig:ssim-ablation}
\end{figure}

\paragraph{Input-side stability.}
Figure~\ref{fig:ssim-ablation} demonstrates input-stability using structural similarity (SSIM) between saliency maps of clean and prediction-preserving Gaussian-noised inputs. Across datasets, natural models are consistently least stable, while adversarial training improves SSIM and smoothing further increases SSIM, especially at moderate-to-high noise levels.
The strongest SSIM gains appear on ImageNette (with Gaussian or Mean filter), where smoothing produces visibly more coherent maps under noise.

\paragraph{Takeaway.} These ablations suggest that \emph{local smoothing} that reduces high-frequency fluctuations in intermediate feature maps is the key mechanism for improved stability rather than Gaussian filtering specifically although the impact differs based on dataset and the filter type. Gaussian remains attractive in practice because it is differentiable, provides smooth frequency attenuation, and yields consistently balanced improvements.
\vspace{-2mm}
\section{Limitations and Future Directions}
Our study focused on input-gradient explanations. It remains unclear whether the same stability trends and training-time fixes extend to other explanation families such as perturbation-based (e.g., LIME~\cite{ribeiro2016should}, SHAP~\cite{lundberg2017unified}) or concept-based methods (e.g., TCAV~\cite{kim2018interpretability}). Second, our experiments are limited to mid-scale vision datasets and convolutional architectures. Evaluating larger-scale settings and modern models (e.g., Transformers and vision--language models) is an important next step, especially since adversarial training can behave differently at scale~\cite{zhang2019limitations}. Finally, we primarily use simple local smoothing (Gaussian, with mean/median ablations). More expressive regularizers~\cite{yoshida2017spectral} or learned smoothing modules may yield better trade-offs between accuracy, robustness, and explanation stability.
\vspace{-3mm}
\section{Conclusion}
We showed that explanation quality is strongly shaped by how a model is trained. While adversarial training typically yields sparser and more input-stable saliency maps, we find it can worsen output-side stability, with explanations changing noticeably under prediction-preserving perturbations. To address this, we augment adversarial training with a lightweight feature-map smoothing block. Across datasets and architectures, smoothing consistently improves stability while largely preserving the sparsity and robustness benefits of adversarial training, and it aligns with higher human ratings of explanation quality. These results support a training-centered path toward more trustworthy gradient-based saliency maps.

{
    \small
    \bibliographystyle{ieeenat_fullname}
    \bibliography{main}
}

\clearpage
\setcounter{page}{1}
\maketitlesupplementary

\appendix 

\section{Training details}\label{appendix:training}

\paragraph{FMNIST \cite{xiao2017fashion}.}
Fashion-MNIST consists of $28\times28$ grayscale images across 10 classes.
Following \cite{chalasani2020concise}, we use a two-layer CNN with 32 and 64 filters, each followed by $2\times2$ max-pooling, and a 1024-unit fully connected layer.
We train with Adam (learning rate $=10^{-3}$), batch size 32, for 50 epochs.

\paragraph{CIFAR-10 \cite{krizhevsky2009learning}.}
CIFAR-10 contains $32\times32$ RGB images in 10 classes.
Following \cite{chalasani2020concise}, we train WideResNet-28-10 \cite{zagoruyko2016wide} for 70k steps with momentum SGD (momentum $=0.9$, weight decay $=5\!\times\!10^{-4}$), batch size 128.
We use a piecewise learning rate schedule: $0.1$ for the first 40k steps, $0.01$ for 40k--50k, and $0.001$ for the remaining steps.
To assess architectural generality, we also report an ablation on VGG \cite{simonyan2014very} trained with the same setup.

\paragraph{ImageNette \cite{howardsmaller}.}
ImageNette is a 10-class subset of ImageNet \cite{deng2009imagenet}.
We use 320-pixel images (short side), apply random resize and crop to $224\times224$ during training, and train ResNet-18 \cite{he2016deep} from scratch.
We use Ranger \cite{Ranger} (learning rate $=8\!\times\!10^{-3}$, $\epsilon=10^{-6}$), train for up to 200 epochs, and select the best checkpoint via early stopping on validation accuracy.

\paragraph{Adversarial training.}
Adversarial training \cite{goodfellow2015explaining} optimizes the min--max objective in Eq.~\eqref{eq:at_obj}.
We approximate the inner maximization using $\ell_\infty$ PGD \cite{madry2017towards}.
For FMNIST and CIFAR-10, we use $\epsilon=0.1$, step size $\alpha=\epsilon/10$, and 40 PGD steps.
For ImageNette, we use $\epsilon=1/255$, step size $\alpha=2/255$, and 20 PGD steps.
All other training hyperparameters match the natural-training setup described above.

\paragraph{Adversarial training with Gaussian smoothing block.}
For \textbf{G}, we insert one SmoothBlock (Sec.~\ref{sec:smoothing}) after the first convolutional/residual block, which we found most effective.
We keep the architecture and training identical to \textbf{A} otherwise.
We also include filter-type ablations (Gaussian vs. mean/median) in Sec.~\ref{ablation:meanMedian} and placement ablations in Sec.~\ref{ablation:position}.

\section{Faithfulness: ROAD plots}
\begin{figure}[h!]
    \centering

    \begin{subfigure}{0.5\textwidth}
        \centering
        \begin{subfigure}[b]{.45\textwidth}
            \centering
            \includegraphics[width=0.9\linewidth]{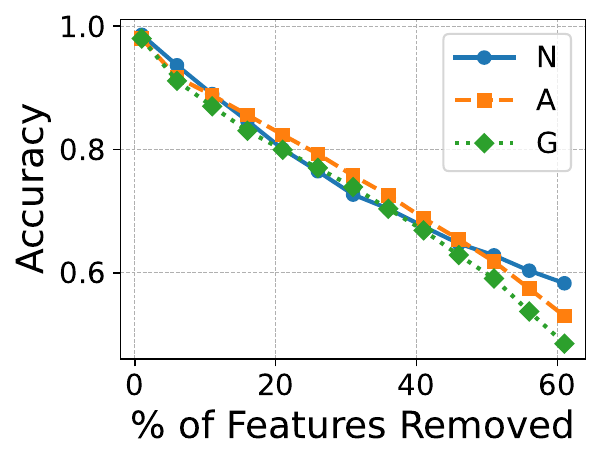}
            \caption{Vanilla Gradient (VG)}
            \label{fig:VGFMNIST_road}
        \end{subfigure}
        \hfill
        \begin{subfigure}[b]{.45\textwidth}
            \centering
            \includegraphics[width=0.9\linewidth]{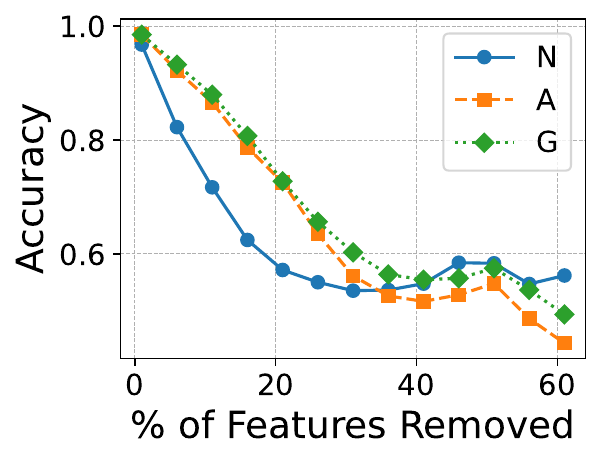}
            \caption{Integrated Gradient (VG)}
            \label{fig:IGFMNIST_road}
        \end{subfigure}
        \caption{FMNIST}
        \label{fig:fmnist_road}
    \end{subfigure}

    \vspace{0.5cm} 

      \begin{subfigure}{0.5\textwidth}
        \centering
        \begin{subfigure}[b]{.45\textwidth}
            \centering
            \includegraphics[width=0.9\linewidth]{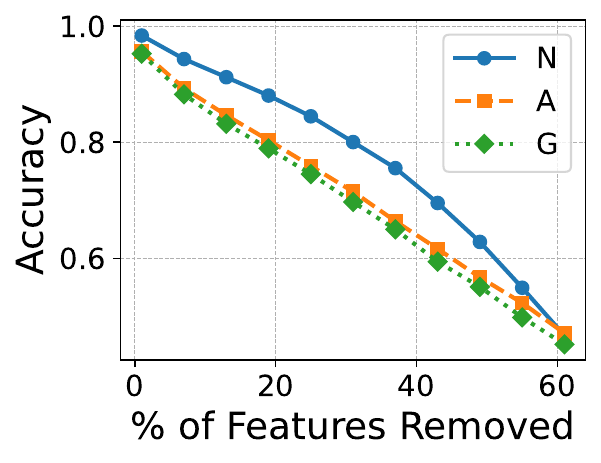}
            \caption{Vanilla Gradient (VG)}
            \label{fig:VGCIFAR_road}
        \end{subfigure}
        \hfill
        \begin{subfigure}[b]{.45\textwidth}
            \centering
            \includegraphics[width=0.9\linewidth]{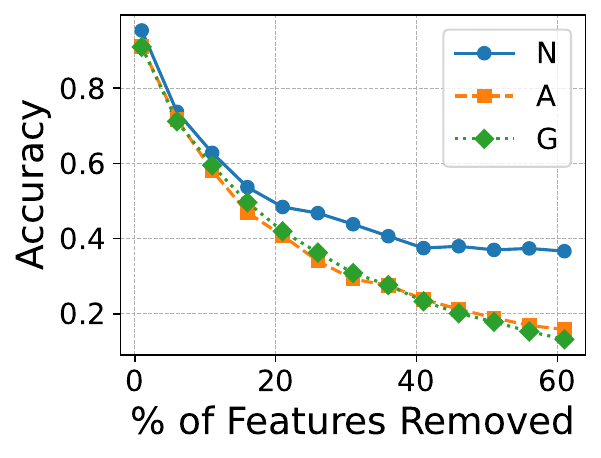}
            \caption{Integrated Gradient (VG)}
            \label{fig:IGCIFAR_road}
        \end{subfigure}
        \caption{CIFAR-10}
        \label{fig:cifar_road}
    \end{subfigure}

      \begin{subfigure}{0.5\textwidth}
        \centering
        \begin{subfigure}[b]{.45\textwidth}
            \centering
            \includegraphics[width=0.9\linewidth]{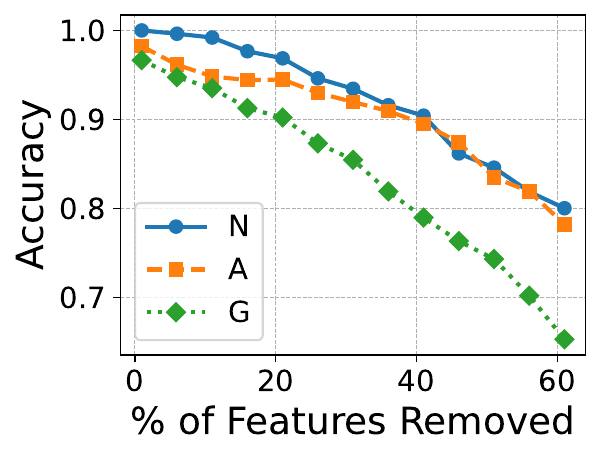}
            \caption{Vanilla Gradient (VG)}
            \label{fig:VGImagenette_road}
        \end{subfigure}
        \hfill
        \begin{subfigure}[b]{.45\textwidth}
            \centering
            \includegraphics[width=0.9\linewidth]{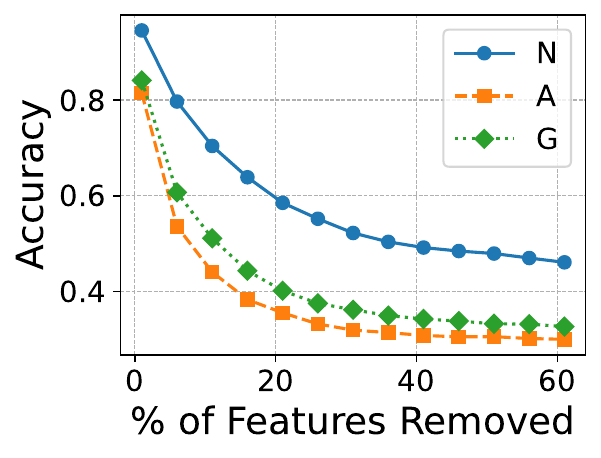}
            \caption{Integrated Gradient (VG)}
            \label{fig:IGImagenette_road}
        \end{subfigure}
        \caption{ImageNette}
        \label{fig:imagenettte_road}
    \end{subfigure}
    \caption{ROAD plots of Vanilla Gradient and Integrated Gradient on FMNIST, CIFAR-10 and ImageNette models: naturally-trained \textbf{(N)}, adversarially trained \textbf{(A)} and adversarially trained with Gaussian filter \textbf{(G)}.}
    \label{fig:combined_road_plot}
\end{figure}

Associated ROAD plots for faithfulness evaluation of different models are provided in Figure~\ref{fig:combined_road_plot}, where sharper drop of accuracy is preferred for faithful explanations.

\section{Closed-form expression for Integrated Gradients (IG)}\label{appendix:closedform}
The feature attribution score computed by Integrated Gradients (IG) for feature $i$ of input image $\mathbf{x} \in R^d$ with baseline $\mathbf{u}$, model $f$ is given by:
\begin{equation}\label{suppeqn:igequationsimple}
    IG_i^f (\mathbf{x,u}) = (x_i - u_i). \int_{\alpha=0}^{1} {\partial_i {f}(\mathbf{u}+\alpha (\mathbf{x} - \mathbf{u}))} \partial \alpha 
\end{equation}

For an input image $\mathbf{x}$, IG returns a vector $IG^F(\mathbf{x}, \mathbf{u}) \in R^d$ with scores that quantify the contribution of $x_i$ to the model prediction ${f}(\mathbf{x})$. For a single layer network ${f}(\mathbf{x}) = H(\langle \mathbf{w},\mathbf{x}\rangle)$ where $H$ is a differentiable scalar-valued function and $\langle \mathbf{w},\mathbf{x}\rangle$ is the dot product between the weight vector $\mathbf{w}$ and input $\mathbf{x}\in R^d$, IG attribution has a closed form expression \cite{chalasani2020concise}.

For given $\mathbf{x}$, $\mathbf{u}$ and $\alpha$, let us consider $\mathbf{v}= \mathbf{u}+\alpha (\mathbf{x}-\mathbf{u})$. If the single-layer network is represented as ${f}(\mathbf{x}) = H(\langle\mathbf{w},\mathbf{x}\rangle)$ where $H$ is a differentiable scalar-valued function, $\partial_i{f}(\mathbf{v})$ can be computed as:

\begin{align}\label{suppeqn:partialfv}
    \partial_i f(\mathbf{v}) &= \frac{\partial \mathcal{F}(\mathbf{v})}{v_i} \notag \\ 
    &= \frac{\partial H(\langle\mathbf{w}, \mathbf{v}\rangle)}{\partial v_i} \notag \\
    &= H'(z) \frac{\partial\langle\mathbf{w}, \mathbf{v}\rangle}{\partial v_i} \notag\\ 
    &= w_i H'(z)  
\end{align}

Here, $H'(z)$ is the gradient of the activation $H(z)$ where $z = \langle\mathbf{w}, \mathbf{v}\rangle$. To compute $\frac{\partial {f}(\mathbf{v})}{\partial \alpha}$:

\begin{equation}\label{suppeqn:computeto}
    \frac{\partial {f}(\mathbf{v})}{\partial \alpha} = \sum_{i=1}^d (\frac{\partial {f}(\mathbf{v})}{\partial v_i} \frac{\partial v_i}{\partial \alpha})
\end{equation}

We can substitute value of $ \frac{\partial v_i}{\partial \alpha} = (x_i-u_i)$ and $\partial_i {f}(\mathbf{v})$ from Eq. \ref{suppeqn:partialfv} to Eq. \ref{suppeqn:computeto}.

\begin{align}
    \frac{\partial {f}(\mathbf{v})}{\partial \alpha} &= \sum_{i=1}^d [w_i H'(z) (x_i - u_i)] \notag \\  
    &= \langle\mathbf{x}-\mathbf{u}, \mathbf{w}\rangle H'(z)
\end{align}

This gives: 
\begin{equation}
    df(\mathbf{v}) = \langle \mathbf{x}-\mathbf{u}, \mathbf{w}\rangle H'(z)\partial\alpha 
\end{equation}

Since $\langle\mathbf{x}-\mathbf{u}, \mathbf{w}\rangle$ is scalar, 
\begin{equation}\label{suppeqn:torewriteIG}
    H'(z)\partial\alpha = \frac{d{f}(\mathbf{v})}{\langle\mathbf{x}-\mathbf{u}, \mathbf{w}\rangle} 
\end{equation}

Eq. \ref{suppeqn:torewriteIG} can be used to rewrite the integral in the definition of $IG_i^f(\mathbf{x})$ in Eq. \ref{suppeqn:igequationsimple},

\begin{align}
    \int_{\alpha=0}^{1} \partial_i f(\mathbf{v})\partial\alpha &= \int_{\alpha=0}^{1} w_i H'(z) \partial z \notag ~~~ \textnormal{[From Eqn. \ref{suppeqn:partialfv}]}\\ 
    &= \int_{\alpha=0}^{1} w_i \frac{ d{f}(\mathbf{v})}{\langle\mathbf{x}-\mathbf{u}, \mathbf{w}\rangle} \notag\\ 
     &= \frac{w_i}{\langle\mathbf{x}-\mathbf{u}, \mathbf{w}\rangle} \int_{\alpha=0}^{1}{d {f}(\mathbf{v})} \notag \\
     &= \frac{w_i}{\langle\mathbf{x}-\mathbf{u}, \mathbf{w}\rangle} [{f}(\mathbf{x})-{f}(\mathbf{u})]
\end{align}

Hence, we obtain the closed form for Integrated Gradients from its definition in Eqn. \ref{suppeqn:igequationsimple} as

\begin{align}\label{suppeqn:igequation}
    IG_i^{f}(\mathbf{x},\mathbf{u}) &= [{f}(\mathbf{x}) - {f}(\mathbf{u})] \frac{({x_i} - {u_i}){w_i}}{\langle\mathbf{x} - \mathbf{u}, \mathbf{w}\rangle} \notag \\
    IG^f(\mathbf{x},\mathbf{u})  &= [{f}(\mathbf{x}) - {f}(\mathbf{u})] \frac{(\mathbf{x} - \mathbf{u}) \odot
 \mathbf{w}}{\langle\mathbf{x} - \mathbf{u}, \mathbf{w}\rangle}
\end{align}

Here, $\odot$ is the entry-wise product of two vectors.

\section{Stability bound for Integrated Gradients}\label{appendix:IGfullproof}

We analyze the single-layer model ${f}(\mathbf{x})=H(\langle \mathbf{w},\mathbf{x}\rangle)$ with a differentiable scalar activation $H$.
For a baseline $\mathbf{u}$, Integrated Gradients (IG) is

\begin{equation}
    \begin{split}
        \phi_{\text{IG}}(\mathbf{x};\mathbf{u})
= (\mathbf{x}-\mathbf{u}) \odot \int_{0}^{1}\nabla_{\mathbf{x}}{f}\!\big(\mathbf{u}+\alpha(\mathbf{x}-\mathbf{u})\big)\,d\alpha
\\ 
= (\mathbf{x}-\mathbf{u}) \odot \left(\int_{0}^{1} H'\!\big(\langle \mathbf{w},\,\mathbf{u}+\alpha(\mathbf{x}-\mathbf{u})\rangle\big)\,d\alpha\right)\mathbf{w}.
    \end{split}
\end{equation}

Equivalently, using the closed form of \cite{chalasani2020concise}, when $\langle \mathbf{w},\mathbf{x}-\mathbf{u}\rangle\neq 0$,

$$
\phi_{\text{IG}}(\mathbf{x};\mathbf{u})
=\big[H(\langle \mathbf{w}, \mathbf{x}\rangle)-H(\langle \mathbf{w}, \mathbf{u}\rangle)\big]\;
\frac{(\mathbf{x}-\mathbf{u})\odot \mathbf{w}}{\langle \mathbf{w}, \mathbf{x}-\mathbf{u}\rangle}.)
$$

Let $\mathbf{d}=\mathbf{x}-\mathbf{u}$ and define the path-average

$$
A(\mathbf{x};\mathbf{u}) \;:=\; \int_{0}^{1} H'\!\big(\langle \mathbf{w},\,\mathbf{u}+\alpha(\mathbf{x}-\mathbf{u})\rangle\big)\,d\alpha.
$$

so that, $ \phi_{\text{IG}}(\mathbf{x};\mathbf{u}) \;=\; \big(\mathbf{d}\odot \mathbf{w}\big)\,A(\mathbf{x};\mathbf{u})$.

Consider a nearby point $\mathbf{x}'$ (same predicted label) and set $\mathbf{d}'=\mathbf{x}'-\mathbf{u}$.
Then, similar to Vanilla Gradient, the stability for Integrated Gradient can be computed as the norm of the difference between IG explanation for noisy image and the original image. 

\begin{equation}
    \begin{split}
        \Delta_{\text{IG}}
\;:=\; \big\|\phi_{\text{IG}}(\mathbf{x}';\mathbf{u})-\phi_{\text{IG}}(\mathbf{x};\mathbf{u})\big\| \\
\;=\; \big\|\;(\mathbf{d}'\!\odot\!\mathbf{w})\,A(\mathbf{x}';\mathbf{u}) \;-\; (\mathbf{d}\!\odot\!\mathbf{w})\,A(\mathbf{x};\mathbf{u})\;\big\|.
    \end{split}
\end{equation}

Because we want to bound the difference of two products (consisting of composite terms), there is no simple rule that lets us directly split this as a product of differences. Unlike sums ($||a+b|| \leq ||a|| + ||b||$, the norm of a difference of products cannot be written as a clean product of differences. So, to separate the two kinds of changes (the change in $d$ and the change in $A$), and apply the triangle inequality, we add–and–subtract $(\mathbf{d}\!\odot\!\mathbf{w})A(\mathbf{x}';\mathbf{u})$):

\begin{equation}
    \begin{split}
      = || \underbrace{(\mathbf{d}'\odot \mathbf{w})A(\mathbf{x}', \mathbf{u}) - (\mathbf{d}\odot \mathbf{w})A(\mathbf{x}', \mathbf{u})}_{\text{difference in }\mathbf{d}} + \\ \underbrace{(\mathbf{d}\odot \mathbf{w})A(\mathbf{x}', \mathbf{u}) - (\mathbf{d}\odot \mathbf{w})A(\mathbf{x}, \mathbf{u})}_{\text{difference in }A}||.
    \end{split}
\end{equation}

Hence, we get,

\begin{equation}
    \begin{split}
        \Delta_{\text{IG}}
\;\le\;
\underbrace{\big\|\mathbf{d}'\!\odot\!\mathbf{w}-\mathbf{d}\!\odot\!\mathbf{w}\big\|}_{\text{(i)}}\;\cdot\; \underbrace{\big|A(\mathbf{x}';\mathbf{u})\big|}_{\text{(ii)}}
\;+\; \\ 
\underbrace{\big\|\mathbf{d}\!\odot\!\mathbf{w}\big\|}_{\text{(iii)}}\;\cdot\; \underbrace{\big|A(\mathbf{x}';\mathbf{u})-A(\mathbf{x};\mathbf{u})\big|}_{\text{(iv)}}.
    \end{split}
\end{equation}

We bound the four factors:

\paragraph{Linear factor in $\mathbf{x}'-\mathbf{x}$} Since 
$\mathbf{d}'-\mathbf{d}=\mathbf{x}'-\mathbf{x}$, so

$$
\big\|\mathbf{d}'\!\odot\!\mathbf{w}-\mathbf{d}\!\odot\!\mathbf{w}\big\|
=\big\|(\mathbf{x}'-\mathbf{x})\!\odot\!\mathbf{w}\big\|
\;\le\; \|\mathbf{w}\|\,\|\mathbf{x}'-\mathbf{x}\|.
$$

\paragraph{Bounding the path-average $A(\cdot)$}  For any $\alpha\in[0,1]$, $|H'(\cdot)|\le \sup_{z}|H'(z)|$. Hence

$$
|A(\mathbf{x}';\mathbf{u})| \;\le\; \sup_{z}|H'(z)|.
$$

\paragraph{Size of $\mathbf{d}\!\odot\!\mathbf{w}$} 
$\big\|\mathbf{d}\!\odot\!\mathbf{w}\big\|\le \|\mathbf{w}\|\,\|\mathbf{d}\|$.



Differentiate inside the integral and apply the chain rule:

$$
\nabla_{\mathbf{x}} A(\mathbf{x};\mathbf{u})
= \int_{0}^{1} \alpha\,H''\!\big(\langle \mathbf{w},\,\mathbf{u}+\alpha(\mathbf{x}-\mathbf{u})\rangle\big)\;\mathbf{w}\;d\alpha,
$$

so

\begin{equation}
    \begin{split}
        \big\|\nabla_{\mathbf{x}} A(\mathbf{x};\mathbf{u})\big\|
\;\le\; \left(\int_{0}^{1}\alpha\,d\alpha\right)\;\sup_{z}|H''(z)|\;\|\mathbf{w}\| \\ 
\;=\;\tfrac12\sup_{z}|H''(z)|\;\|\mathbf{w}\|.
    \end{split}
\end{equation}

By the mean-value theorem along the line segment $[\mathbf{x},\mathbf{x}']$,
$$
\big|A(\mathbf{x}';\mathbf{u})-A(\mathbf{x};\mathbf{u})\big|
\;\le\; \tfrac12\sup_{z}|H''(z)|\;\|\mathbf{w}\|\;\|\mathbf{x}'-\mathbf{x}\|.
$$

Putting the bounds together:
\begin{equation}
    \begin{split}
        \Delta_{\text{IG}}
\;\le\;
\underbrace{\|\mathbf{w}\|\,\|\mathbf{x}'-\mathbf{x}\|}_{(i)}\cdot \underbrace{\sup_{z}|H'(z)|}_{(ii)}
\;+\; \\
\underbrace{\|\mathbf{w}\|\,\|\mathbf{d}\|}_{(iii)}\cdot
\underbrace{\left(\tfrac12\sup_{z}|H''(z)|\;\|\mathbf{w}\|\;\|\mathbf{x}'-\mathbf{x}\|\right)}_{(iv)}
    \end{split}
\end{equation}

Equivalently,

$$
\Delta_{\text{IG}} \leq \;\Big(\sup_{z}|H'(z)|\,\|\mathbf{w}\| \;+\; \tfrac12\sup_{z}|H''(z)|\,\|\mathbf{w}\|^2\,\|\mathbf{d}\|\Big)\;\cdot\;\|\mathbf{x}'-\mathbf{x}\|.
$$

Collect terms into a single constant that depends only on local quantities:

\begin{equation}
    \begin{split}
        \Delta_{\mathrm{IG}}
\;\le\;
C_{\mathrm{IG}}\;\|\mathbf{x}'-\mathbf{x}\|,
\quad \\ 
\text{with} \quad
C_{\mathrm{IG}}
:=
\sup_{z}|H'(z)|\,\|\mathbf{w}\|
\;+\; \\ 
\tfrac12\,\sup_{z}|H''(z)|\;\|\mathbf{w}\|^2\|\,\|\mathbf{x}-\mathbf{u}\|.
    \end{split}
\end{equation}

Thus, for the single-layer model, IG stability—like VG—is primarily controlled by activation curvature $\sup|H''|$ and the weight scale $\|\mathbf{w}\|$.

\section{Additional architecture: VGG16}\label{appendix:additionalexperiments}
To assess whether our findings generalize beyond LeNet \cite{lecun1998gradient} and residual networks \cite{he2016deep}, we repeat our study with a VGG-16 architecture \cite{simonyan2014very} on CIFAR-10. We apply natural (N), adversarial (A) and adversarial training with Gaussian filter (G) following the same setup as explained in Sec.~\ref{sec:experimentAndAnalysis}.

\begin{table}[h]
\centering
\caption{\textbf{Ablation:} Sparsity and stability results for VGG16 on CIFAR-10 models: naturally trained \textbf{(N)}, adversarially trained \textbf{(A)}, \& adversarially trained with Gaussian smoothing \textbf{(G)}}
\label{tab:vgg16}
\resizebox{0.30\textwidth}{!}{%
\begin{tabular}{@{}llccc@{}}
\toprule
            &             & \textbf{N} & \textbf{A} & \textbf{G} \\ \midrule
\textbf{VG} & Gini $\uparrow$ & 0.45 & 0.55 & 0.55 \\
            & ROS $\downarrow$  & 2.34 & 2.04 & 2.01 \\ \midrule
\textbf{IG} & Gini $\uparrow$ & 0.61 & 0.64 & 0.63 \\
            & ROS $\downarrow$  & 4.23 & 4.34 & 4.03 \\ \bottomrule
\end{tabular}%
}
\end{table}

\noindent\textbf{Results.} The results in Table~\ref{tab:vgg16} mirror our ResNet findings from Table~\ref{tab:sparsity-stability}: adversarial training improves sparsity (higher Gini), while Gaussian smoothing restores or further improves stability (ROS) with minimal change in sparsity. Although sparsity gains are less pronounced in VGG16 compared to ResNet, stability benefits are consistent, supporting the generality of our approach across architectures.

\section{User-study}\label{appendix:userstudy}

To ensure that the observed differences in participant ratings were statistically meaningful rather than random variations, we performed Wilcoxon signed-rank tests \cite{woolson2007wilcoxon} and one-way ANOVA \cite{cuevas2004anova}. The Wilcoxon test assesses whether paired differences between two conditions (e.g., adversarially trained vs. naturally trained) are statistically significant, making it well-suited for analyzing subjective survey responses. The one-way ANOVA test determines whether there are significant differences across all three models. As shown in Table \ref{tab:stattest}, the extremely small p-values ($<$ 0.001) indicate that differences in both sufficiency and trust scores across training strategies are statistically significant. 

\begin{table}[h]
\centering
\caption{Wilcoxon and ANOVA test results on the survey where, N refers to a naturally trained model, A refers to an adversarially-trained model, and G refers to an adversarial-trained feature-map smoothed model.}
\label{tab:stattest}
\resizebox{0.49\textwidth}{!}{%
\begin{tabular}{lccccc}
\toprule
 & \multicolumn{3}{c}{\textbf{Wilcoxon signed-rank}} & \multicolumn{2}{c}{\textbf{(Reported) ANOVA}} \\
\cmidrule(lr){2-4}\cmidrule(lr){5-6}
 & \textbf{N vs A} & \textbf{A vs G} & \textbf{N vs G} & \textbf{F-stat} & \textbf{p-value} \\
\midrule
\textbf{Sufficiency} & 9.79E-41 & 4.26E-14 & 3.71E-27 & 200.38 & 7.82E-72 \\
\textbf{Trust}       & 5.56E-39 & 3.24E-11 & 3.89E-24 & 193.86 & 6.58E-70 \\
\bottomrule
\end{tabular}%
}
\end{table}

\begin{figure}[h]
    \centering
    \includegraphics[width=0.95\linewidth]{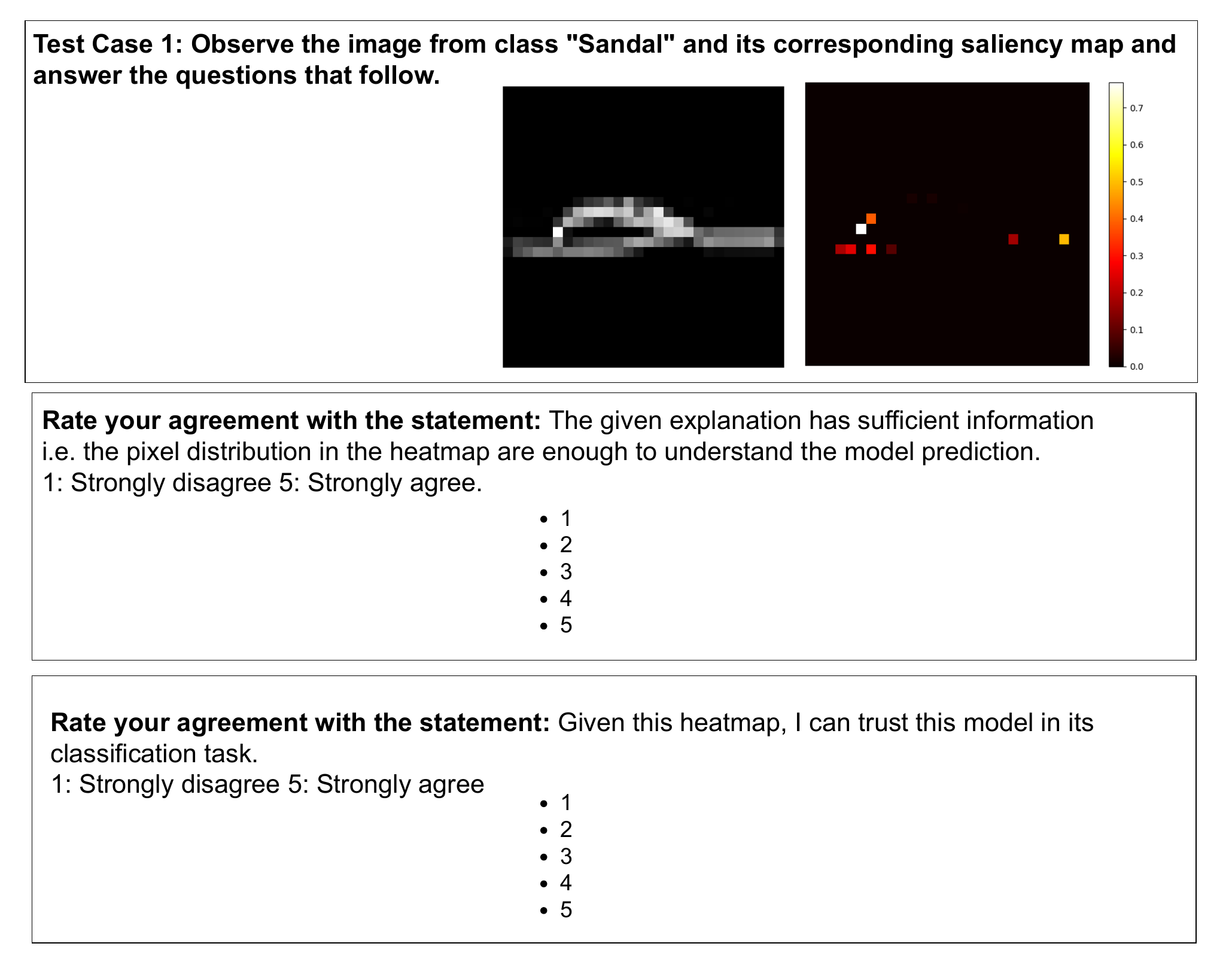}
    \caption{A sample of question from the survey}
    \label{fig:survey1}
\end{figure}

\begin{figure}[h]
    \centering
    \includegraphics[width=0.42\linewidth]{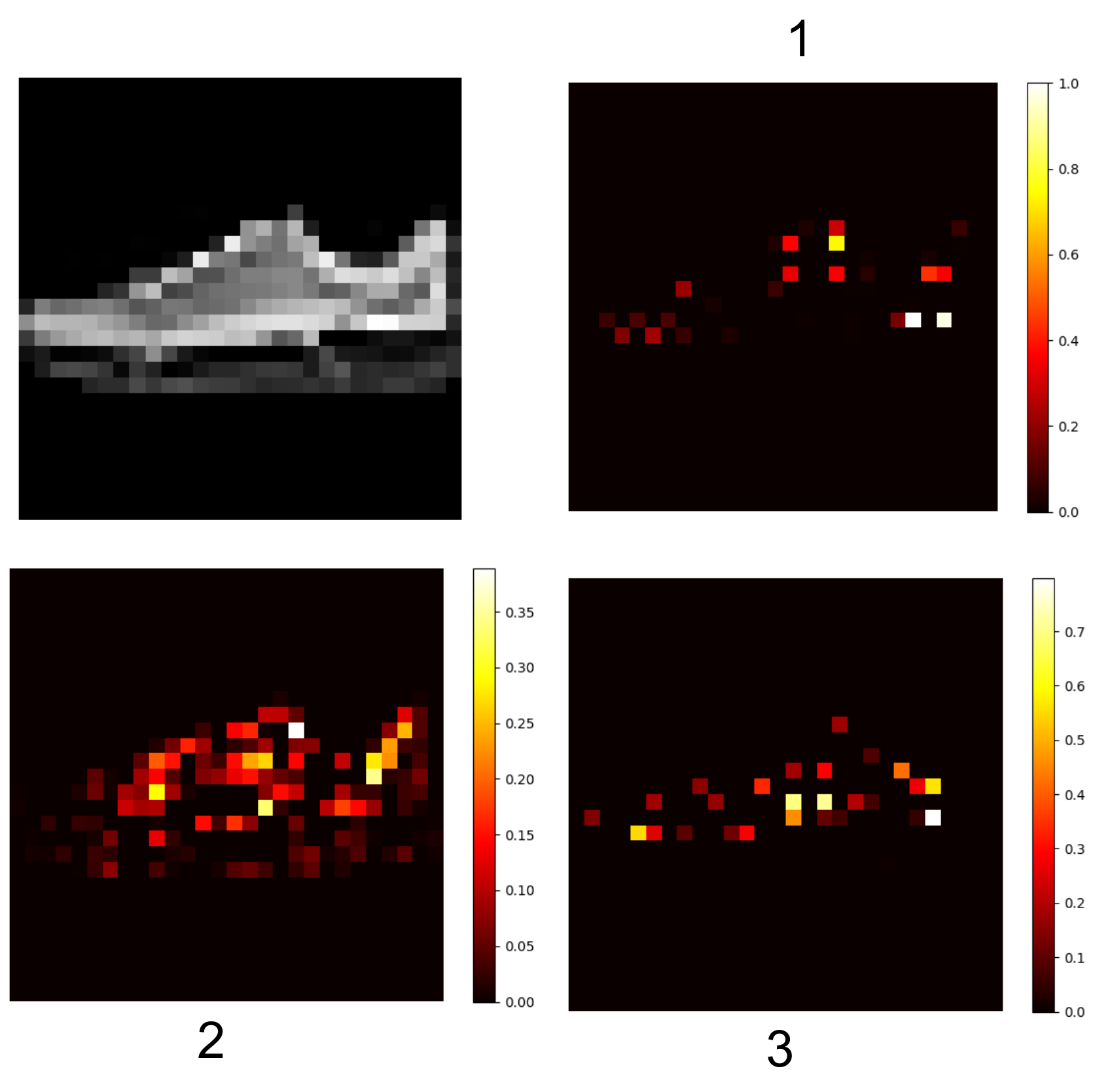}
    \caption{Final block sample images in the survey}
    \label{fig:survey2}
\end{figure}

\section{Evaluating explanations}\label{appendix:metrics}
Evaluating saliency maps remains challenging, as qualitative visual assessments can be subjective. Recent research emphasizes the need for functionally grounded metrics.  

\textbf{Sparsity.} Chalasani et al.~\cite{chalasani2020concise} proposed the Gini index to quantify sparsity of attribution vectors, encouraging explanations that focus on a small set of discriminative features.  Given a vector of attribution $\phi(\mathbf{x}) \in R^d$, the absolute of the vector is first sorted in non-decreasing order, and the Gini index is computed using Eqn. \ref{eqn:gini}. 

\begin{equation}\label{eqn:gini}
    G(\phi(\mathbf{\mathbf{x}})) = 1 - 2 \sum_{k=1}^d \frac{\phi(\mathbf{x})_{(k)}}{||\phi(\mathbf{x})||_1} \frac{d-k+0.5}{d}
\end{equation}
 
The Gini Index values lie in between $[0,1]$; A value of 1 indicates perfect sparsity, where only one element in the vector $\phi_i(\mathbf{x}) > 0$. The sparsity is zero if all the vectors are equal to some positive value. 

\textbf{Stability.} \textit{Stability} in explanations quantifies the variance of attribution vectors within a small neighborhood of a given input. For similar inputs, it is expected that the explanations remain similar. We measure two type of stability. 

For \emph{input-side  stability}, we use structural similarity index (SSIM). For each image, we add Gaussian noise and generate its noisy version such that the model prediction is consistent. We then compute the saliency map of the two images and measure the structural similarity between the maps following Adebayo et al.\cite{adebayo2018sanity}. Higher SSIM indicates structurally consistent attributions.

For \emph{output-side stability}, we following Agarwal et al. \cite{agarwal2022rethinking}, and measure relative output stability (ROS) which, which relates changes in attribution to changes in logits:

\begin{equation}\label{eqn:ros}
\begin{split}
ROS = max_{\mathbf{x}'} \frac{||\frac{\phi(\mathbf{x})-\phi(\mathbf{x}')}{\phi(\mathbf{x})}||}{max(||{z(\mathbf{x})-z(\mathbf{x}')}||_p, \epsilon_{min})} \\ 
\forall \mathbf{x}'~s.t.~\mathbf{x}' \in \mathcal{N}_\mathbf{x}; \hat{y}_\mathbf{x} = \hat{y}_{\mathbf{x}'}
\end{split}
\end{equation}

where $z(\cdot)$ are logits, and $\mathcal{N}_{\mathbf{x}}$ is an $\ell_\infty$ ball with consistent model predictions. Lower ROS is better (more stable for similar predictions).

\textbf{Faithfulness.} Faithfulness measures whether explanations reflect the model’s actual decision-making. Early approaches such as insertion/deletion~\cite{petsiukrise} and ROAR~\cite{hooker2019benchmark} either suffer from distribution shift or are computationally expensive. ROAD~\cite{rong22consistent} mitigates these issues by combining feature removal with debiasing, producing a stable, efficient metric.

ROAD measures the accuracy of a model on the provided test set at each step of an iterative process of removing $k$ most important pixels. Removal of pixels is done with a noisy linear imputation to avoid out-of-distribution samples. We set $k=5$ in our experiments, and adopt the MoRF (Most Relevant First) removal strategy where a faster drop in accuracy with increase in removal of $k$ most important features indicate that key discriminative features are being removed. ROAD demonstrates consistent results with both MoRF and LeRF (Least Removal First) removal strategy. 

We quantify ROAD plot using ROAD-AOPC~score:
\begin{equation}
\mathrm{AOPC} = \frac{1}{L+1} \sum_{k=1}^L \Big(f(\mathbf{x}^{(0)}) - f(\mathbf{x}^{(k)})\Big),
\end{equation}

where $\mathbf{x}^{(k)}$ is obtained by removing the $k$ most relevant pixels and $f(\cdot)$ denotes the model output. Higher AOPC indicates that the explanation better identifies features essential for prediction, i.e., more faithful attributions.

\section{Implementation details: Filters}\label{appendix:filters}
\begin{itemize}
    \item \textbf{Mean filter:} A mean filter replaces each feature with the average of nearby features within a defined kernel. For an input feature map ($I$) of size $H$x$W$ and a $K$-sized kernel, the output feature map $O(u,v)$ is calculated using Eqn. \ref{eqn:mean}:
 \begin{equation}\label{eqn:mean}
     O(u,v) = \frac{1}{K^2} \sum_{i=0}^{K-1}\sum_{j=0}^{K-1}I(u+i, v+j)
 \end{equation}

Here, $u$ and $v$ represent spatial coordinates in the output feature map, ranging from 0 to $H-K$ and 0 to $W-K$ respectively. $I(u+i, v+j)$ denotes the feature value at spatial location $(u+i, v+j)$ in the input feature map. This operation is applied independently to each channel of the input.

\item 
 \textbf{Median filter:} A median filter computes the median value within a small sliding window over the feature map, given by Eqn. \ref{eqn:median}. Given an input feature map $I$ and a median filter window size $K$, the output feature map $O(u,v)$ is computed using Eqn. \ref{eqn:median}:

\begin{equation}\label{eqn:median}
     O(u,v) = median(I(u-\frac{K}{2}:u+\frac{K}{2}, v-\frac{K}{2}:v+\frac{K}{2})
 \end{equation}

Here, $I(u-\frac{K}{2}:u+\frac{K}{2}, v-\frac{K}{2}:v+\frac{K}{2})$ represents the subset of the input feature around $(u,v)$ with a size of $K$x$K$. This operation is applied independently to each channel of the input feature map.

\item \textbf{Gaussian filter:} A Gaussian filter applies a smoothing effect to feature maps by convolving them with a Gaussian kernel, effectively reducing Gaussian noise. The degree of smoothing can be adjusted by modifying the standard deviation (\( \sigma \)) of the Gaussian kernel. Given an input feature map $I$ and a Gaussian filter kernel $K$, the output feature map $O(u,v)$ is calculated with Eqn. \ref{eqn:gaussian}:

 \begin{equation}\label{eqn:gaussian}
     O(u,v) = (I*K)(u,v)
 \end{equation}

Here, $*$ denotes 2D convolution. The Gaussian kernel $K$ is generated using a Gaussian function with a specific standard deviation $\sigma$, defined in Eqn. \ref{eqn:kernelgaus}:

 \begin{equation}\label{eqn:kernelgaus}
     K(u,v) = \frac{1}{2\pi \sigma^2} e^{(-\frac{u^2+v^2}{2\sigma^2})}
 \end{equation}
This operation is independently applied to each channel of the input feature map.
\end{itemize}

\textbf{Implementation:} We utilize the differentiable local filters available in Kornia \cite{eriba2020kornia}. We use a 3x3 Kernel for mean, median, and Gaussian filtering. The standard deviation of the kernel for Gaussian filtering was computed as $(0.3 * ((\mathbf{x}.shape[3] - 1) * 0.5 - 1) + 0.8, 0.3 * ((\mathbf{x}.shape[2] - 1) * 0.5 - 1) + 0.8)$ where $\mathbf{x}$ is the input image.

\section{Additional visualization}\label{appendix:vizvisualizations}

\begin{figure}[h!]
    \centering
    \includegraphics[width=0.45\textwidth]{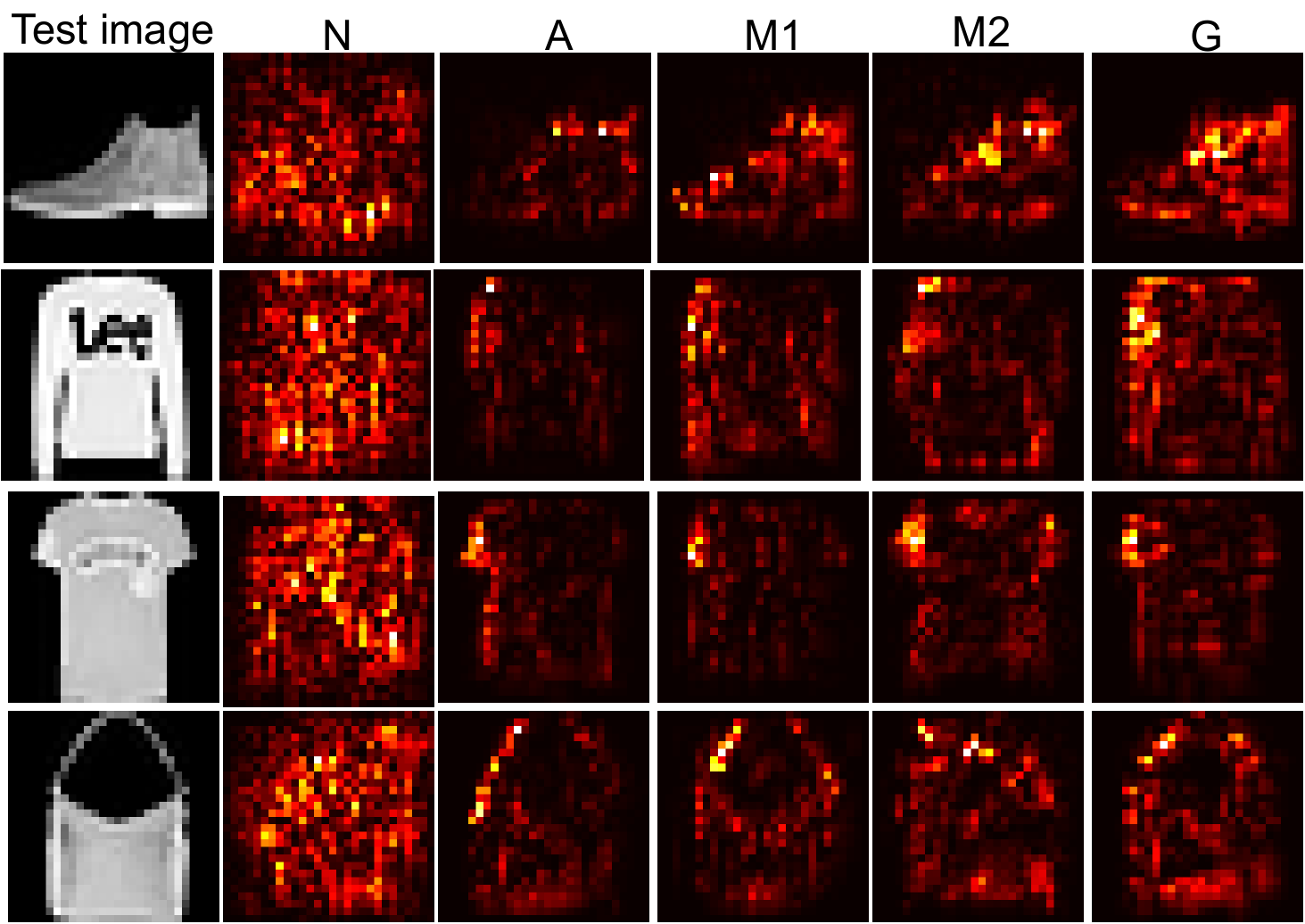}
    \caption{Additional visualization for VG (FMNIST) (N: naturally-trained, A: adversarially-trained, M1: adversarially-trained with mean-filter, M2: adversarially-trained with median-filter, G: adversarially-trained with Gaussian-filter)}
    \label{fig:suppfmnist}
\end{figure}

\begin{figure}[h]
    \centering
    \includegraphics[width=0.45\textwidth]{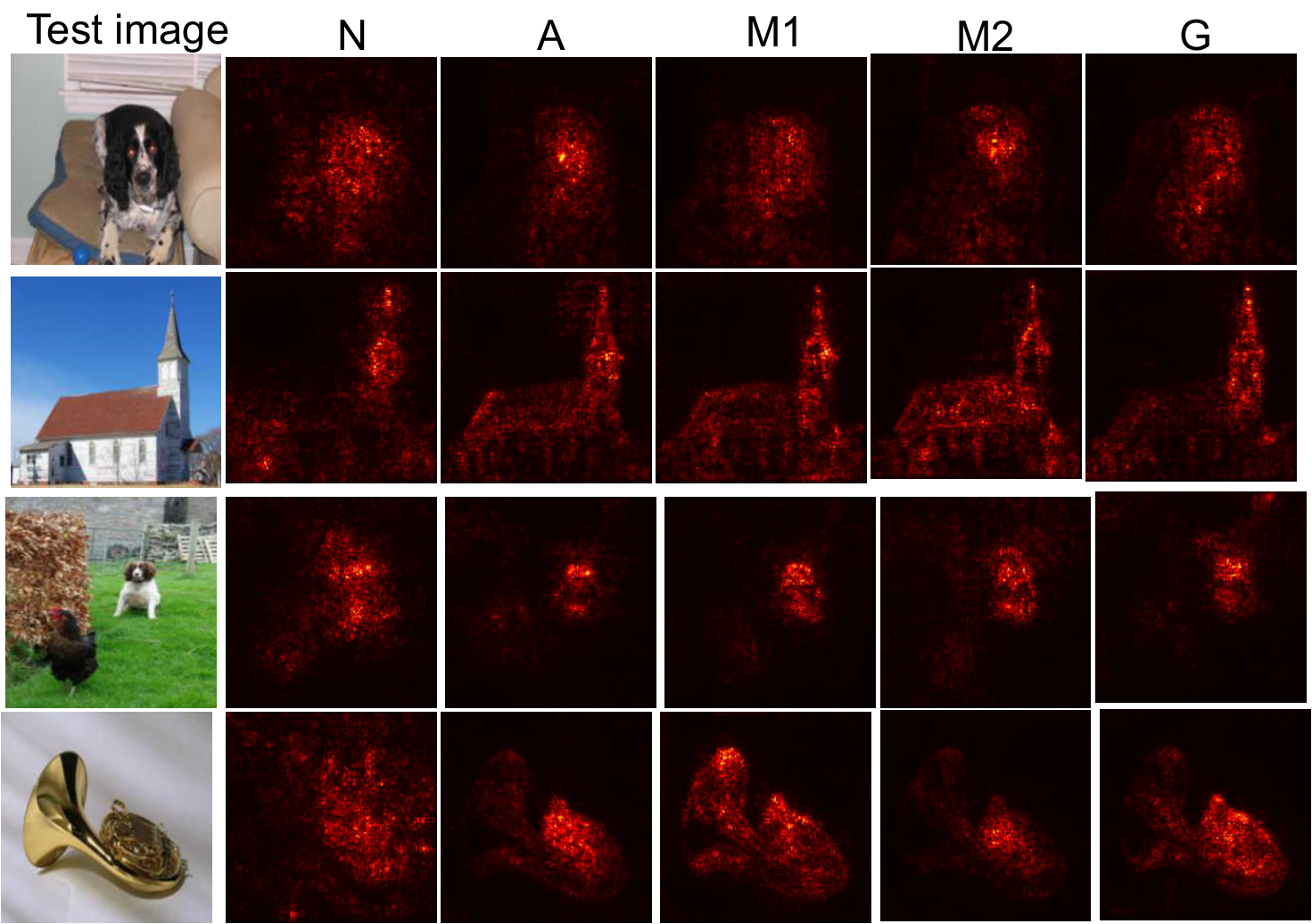}
    \caption{Additional visualization for VG (ImageNette) (N: naturally-trained, A: adversarially-trained, M1: adversarially-trained with mean-filter, M2: adversarially-trained with median-filter, G: adversarially-trained with Gaussian-filter, E: adversarially-trained with embedded filter, NG: adversarially-trained with non-local gaussian)}
    \label{fig:suppimagenette}
\end{figure}

\begin{figure}[h]
    \centering
    \includegraphics[width=0.45\textwidth]{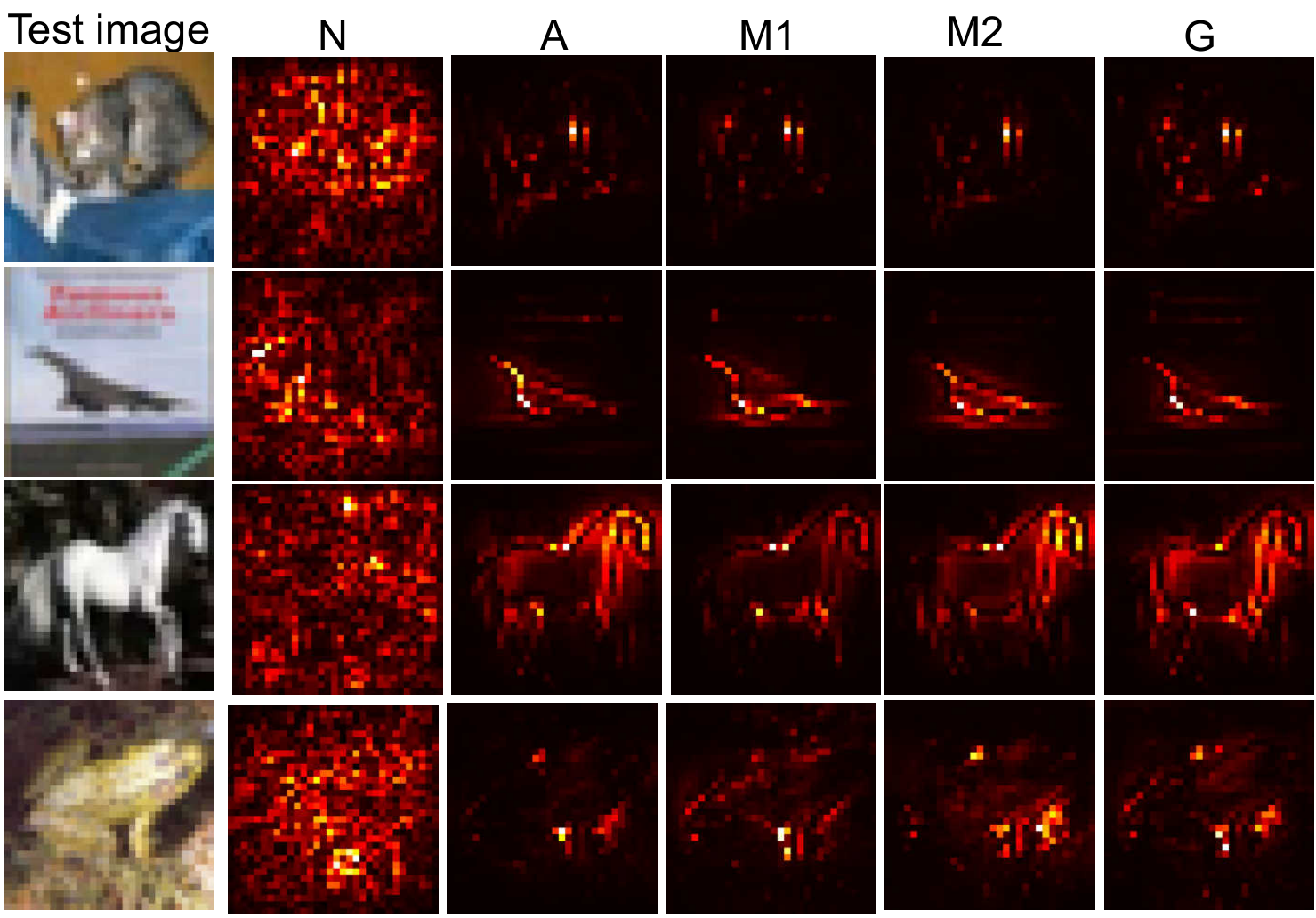}
    \caption{Additional visualization for VG (CIFAR-10) (N: naturally-trained, A: adversarially-trained, M1: adversarially-trained with mean-filter, M2: adversarially-trained with median-filter, G: adversarially-trained with Gaussian-filter, E: adversarially-trained with embedded filter, NG: adversarially-trained with non-local gaussian)}
    \label{fig:suppcifar}
\end{figure}

We provide additional visualizations on Vanilla Gradient (VG) in Figures \ref{fig:suppfmnist}, \ref{fig:suppimagenette} and \ref{fig:suppcifar} for various models: naturally-trained (N), adversarially-trained (A), adversarial training with mean-filter smoothing (M1), adversarial training with median-filter smoothing (M2), adversarial training with Gaussian-filter smoothing (G).

\end{document}